\documentclass[twoside]{article}      
\usepackage{amsmath,amssymb,amsfonts} 
\usepackage{graphics}                 
\usepackage{color}                    
\usepackage{hyperref}                 
\usepackage{comment} 			
\usepackage{enumitem}

\newtheorem{theorem}{Theorem}[section]

\newtheorem{corollary}{Corollary}[theorem]

\newtheorem{definition}[theorem]{Definition}

\date{\small\it \today}
\title{Descriptions of Objectives and Processes of \\ Mechanical Learning
\footnote{Great thanks for whole heart support of my wife. Thanks for Internet and research contents contributers to Internet.}}

\author{ Chuyu Xiong \\
{\small Independent researcher, New York, USA} \\
{\small Email: chuyux99@gmail.com}
}

\begin{document}
\maketitle
\begin{abstract}
In  \cite{paper1}, we introduced mechanical learning and proposed 2 approaches to mechanical learning.  Here, we follow one such approach to well describe the objects and the processes of learning. We discuss 2 kinds of patterns: objective and subjective pattern. Subjective pattern is crucial for learning machine. We prove that for any objective pattern we can find a proper subjective pattern based upon least base patterns to express the objective pattern well. X-form is algebraic expression for subjective pattern. Collection of X-forms form internal representation space, which is center of learning machine. We discuss learning by teaching and without teaching. We define data sufficiency by X-form. We then discussed some learning strategies. We show, in each strategy, with sufficient data, and with certain capabilities, learning machine indeed can learn any pattern (universal learning machine). In appendix, with knowledge of learning machine, we try to view deep learning from a different angle, i.e. its internal representation space and its learning dynamics.

\end{abstract}

 \\

If you want to know the taste of a pear, you must change the pear by eating it yourself. ...... \\
All genuine knowledge originates in direct experience. \\
\indent \hspace{20pt} ----Mao Zedong \\

But, though all our knowledge begins with experience, it by no means follows that \\
all arises out of experience. \\
\indent \hspace{20pt} ----Immanuel Kant \\

Our problem, ...... is to explain how the transition is made from a lower level of knowledge \\
to a level that is judged to be higher. \\
\indent \hspace{20pt} ----Jean Piaget

\section{Introduction}
Mechanical learning is a computing system that is based on a simple set of fixed rules (so called mechanical), and can modify itself according to incoming data (so called learning). A learning machine $\mathcal{M}$ is a system that realizes mechanical learning. 

In \cite{paper1}, we introduced mechanical learning and discussed some basic aspects of it. Here, we are going to continue the discussion of mechanical learning. As we proposed in \cite{paper1}, there are naturally 2 ways to go: to directly realize one learning machine, or to well describe what mechanical learning is really doing. Here, we do not try to design a specific learning machine, instead, we focus on describing the mechanical learning, specially, the objects and the process of learning, and related properties. Again, the most important assumption is mechanical, i.e., the system must follow {\it a set of simple and fixed rules}. By posting such requirement on learning, we can go deeper and reveal more interesting properties of learning.   

In section 2, we discuss more about learning machine. We show one useful simplification: a $N$-$M$ learning machine can be reduced to $M$ independent $N$-1 learning machines. This simplification could help us a lot. We define level 1 learning machine in section 2. This concept clarifies a lot of confusing.

The driving force of a learning machine $\mathcal{M}$ is its incoming data, and incoming data forms patterns. Thus, we need to understand pattern first. In section 3, we discuss patterns and examples. In the process of understanding pattern, what is objective and what is subjective is naturally raised. In fact, these issues are very crucial to learning machine. Objective patterns and their basic operators are straightforward. In order to understand subjective pattern, we discuss how learning machine to perceive and process pattern. Such discussions lead us to subjective pattern and basic operators on them. We introduce X-form for subjective expressions, which will play central role in our later discussions. We prove that for any objective pattern we can find a proper X-form based upon least base patterns and to express the objective pattern well. 

Learning by teaching, i.e. learning driving by a well designed teaching sequence (a special kind of data sequence), is a much simpler and effective learning. Though learning by teaching is only available in very rare cases, it is very educational to discuss it first. This is what we do in section 4. We show if a learning machine has certain capabilities, we can make teaching sequence so that under driven of such teaching sequence, it learns effectively. So, with these capabilities, we have an {\it universal learning machine}.

From learning by teaching, we get insight that the most crucial part of learning is abstraction from lower to higher. We try to apply such insights to learning without teaching. In section 5, we first defined mechanical learning without teaching. Then we introduce {\it internal representation space}, which is the center of learning machine and best to be expressed by X-forms. Internal representation space is actually where learning is happening. We write down the formulation of learning dynamics, which gives a clear picture about how data drives learning. However, one big issue is how much data are enough to drive the learning to reach the target. With the help of X-form and sub-form, we define data sufficiency: sufficient to support a X-form, and sufficient to bound a X-form. Such sufficiency gives a framework for us to understand data used to drive learning. We then show that by a proper learning strategy, with sufficient data, with certain learning capabilities, a learning machine indeed  can learn. We demonstrate 3 learning strategies: embed into parameter space, squeezing to higher abstraction from inside, and squeezing to higher abstraction from inside and outside. We show that the first learning strategy is actually what deep learning is using (see Appendix for details).  And, we show that by other 2 learning strategies with certain learning capabilities, a learning machine can learn any pattern, i.e. it is an {\it universal learning machine}. Squeezing to higher abstraction and more generalization is one strategy that we invent here. We believe that this strategy would work well for many learning tasks. We need to do more works in this direction.

In Section 6, we put more thoughts about learning machine. We will continue work on these directions. In section 7, we briefly discuss some issues of designing a learning machine. 

In Appendix, we view deep learning (restricted to the stacks of RBMs) from our point of view, i.e. internal representation space. We start discussions from simplest, i.e. 2-1 RBM, then 3-1 RBM, N-1 RBM, N-M RBM, and stacks of RBM, and deep learning. In this way, it is clear that deep learning is using the learning strategy: embed a group of X-forms into parameter space that we discuss in section 5. 

As in \cite{paper1}, for the same reason, here we will restrict to {\it spatial learning}, not consider {it temporal learning}.

\section{Learning Machine}\label{table}
{\bf IPU -- Information Processing Unit}\\
We have discussed mechanical learning in \cite{paper1}. A learning machine is a concrete realization of mechanical learning. We can briefly recall them here. See the illustration of IPU (Information Processing Unit):

\begin{center}
\begin{picture}(300,150)(0,0)
\put(0,-200){\resizebox{10 cm}{!}{\includegraphics{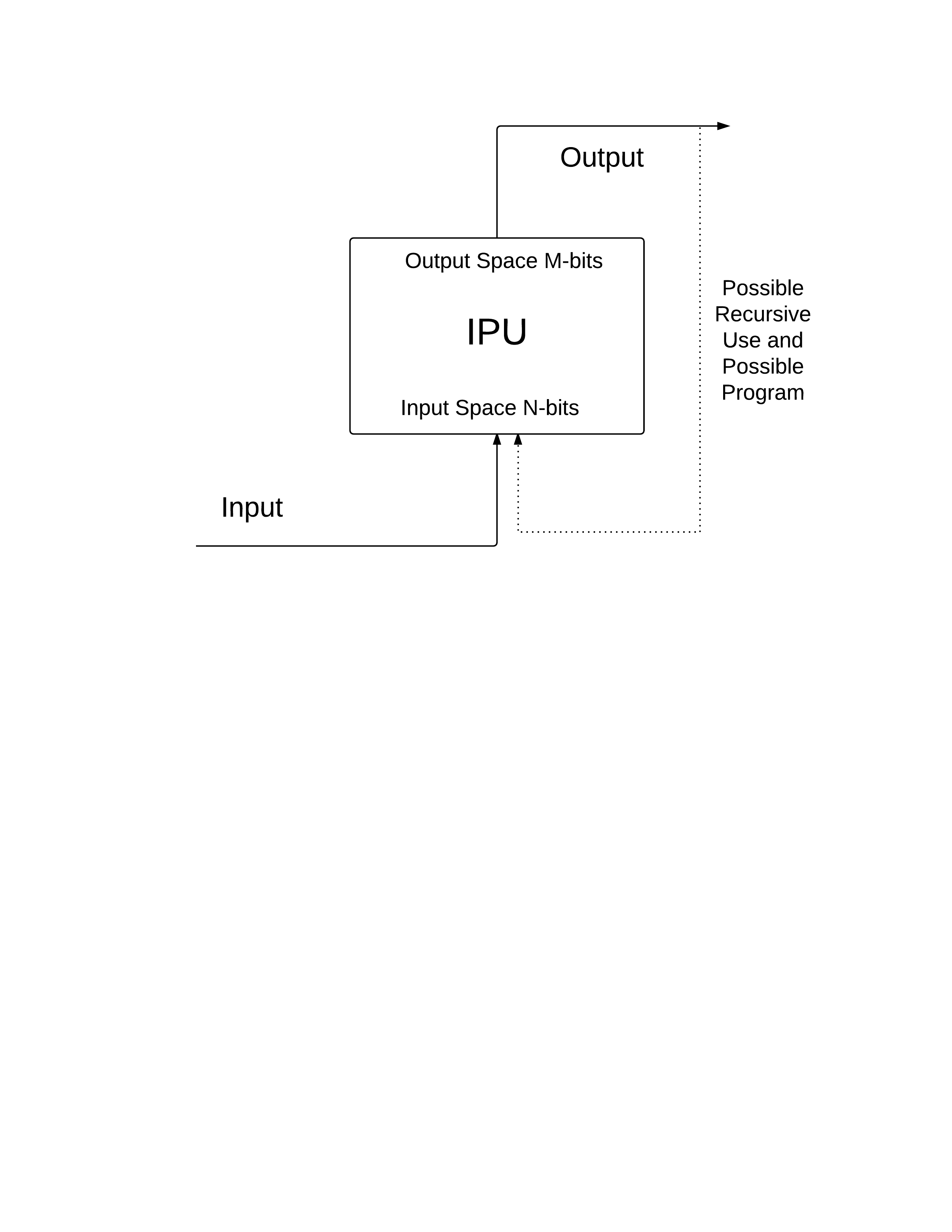}}}
\end{picture}

{\bf Fig. 1. Illustration of $N$-$M$ IPU (Information Processing Unit)} 
\end{center}

One $N$-$M$ IPU has input space ($N$ bits) and output space ($M$ bits), and it will process input to output. If the processing are adapting according to input and feedback to output, and such adapting is governed by a set of simple and fixed rules, we call such adapting as {\it mechanical learning}, and such IPU as {\it learning machine}. Notice the phrase "a set of simple and fixed rules". This is a strong restriction. Mostly, we use this phrase to rule out human intervention. And, we pointed out this: since the set of adapting rules is fixed, we can reasonablly think the adapting rules are built inside learning machine at the setup. 

We will try to well describe learning machine. First, we can put one simple observation here.

\begin{theorem}
One $N$-$M$ IPU $\mathcal{M}$ is equivalent to $M$ $N$-1 IPU $\mathcal{M}_i,  i=1, \ldots, M$.
\end{theorem}
{\bf Proof:} The output space of $\mathcal{M}$ is $M$-dim, so, we assume it is $(v_1, v_2, \ldots, v_M)$. If we project to first component, i.e. $v_1$, we get a $N$-1 IPU, denote it as: $\mathcal{M}_1$. We can do same for $v_i, i = 2, \ldots, M$, and get $N$-1 IPUs: $\mathcal{M}_2, \ldots, \mathcal{M}_M$. This tells us, if we have one $N$-$M$ IPU $\mathcal{M}$, we can get $M$ $N$-1 IPU $\mathcal{M}_1, \ldots$, so that $\mathcal{M}$ = $(\mathcal{M}_1, \mathcal{M}_2, \ldots  \mathcal{M}_M)$.

On the other side, if we have $M$ $N$-1 IPU, $\mathcal{M}_1, \mathcal{M}_2, \ldots,  \mathcal{M}_M$, we can use them to form a $N$-$M$ IPU in this way:  
$\mathcal{M}$ = $(\mathcal{M}_1, \mathcal{M}_2, \ldots  \mathcal{M}_M)$.

Though this theorem is very simple, it can make our discussion much simpler. For most time, we can only consider $N$-1 IPU, which is much simpler to discuss. However, this is only to consider IPU, i.e. ability to process information. For learning, we need to consider more. See theorem 2. 
\bigskip

{\bf The purpose or target of learning machine:} \\
One learning machine is one IPU, i.e. it will do information processing for each input and
generate output, it maps one input (a $N$-dim binary vector) to a $M$-dim binary vector. This is what a CPU does as well (More abstractly, since we do not restrict the size of $N$ and $M$, any software without temporal effect can be thought as one IPU). 

However, learning machine and CPU have very different goal. One CPU is designed to distinguish a
input $b \in PS^0_N$ from any other, even there is only one bit difference, i.e. bit-wise. Yet, IPU and learning machine are not designed for such purpose. IPU and learning machine are designed to distinguish patterns. It should generate different output for different patterns, but, should generate same output for different inputs of a same pattern. That is to say, the target of a learning machine is to learn to distinguish a group of base patterns and how to process them. Thus, we need to understand patterns. Actually, to understand patterns is the most essential job, which is done in next section. 
\bigskip

{\bf Data } \\
The purpose of a learning machine is to learn, i.e. to modify its information processing. However, we would emphasis that for mechanical learning, learning is driven by data fed into it.

\begin{definition}[\bf Data Sequence]
If we have a sequence $T_i, i=1, 2, \ldots$, and $T_i = (b_i, o_i)$, where $b_i$ is a base pattern, $o_i$ is either $\varnothing$ (empty) or a binary vector in output space, we call this sequence a data sequence. 
\end{definition}
Note, $o_i$ could be empty or a vector in output space. If it is non-empty, it means that at the moment, the vector should be the value of output. If it is empty, it means there is no data for output match up. Learning machine should be able to learn even $o_i$ is empty. Of course, with value of output, the learning is often easier and faster.  

We can easily see that data sequence is the only information source for a learning machine to modify itself. Without information from data sequence, learning machine just has no information about what to modify. Learning machine will adapt itself only based on information from data sequence. 

There are 2 kinds of data sequence. One is very well designed data sequence, i.e. we know the consequence of this data, and we can expect the outcome of learning. This is called teaching sequence. Another kind of data sequence is not teaching sequence. These data sequences are just outside data to drive the learning machine (could be random from outside). We have no much knowledge about them. Clearly, in order to learn certain target, if available, a teaching sequence is much more efficient. However, in most cases, we just do not have teaching sequence. 
\bigskip

{\bf Universal Learning Machine } \\
Naturally, we will ask what a learning machine can learn? Can it learn anything? To address this, we need some careful defintion. Suppose we have a learning machine $\mathcal{M}$. At the beginning, $\mathcal{M}$ has the processing $P_0$, i.e. $P_0$ is one mapping from input space ($N$-dim) to output space ($M$-dim). As the learning going, the processing will changed to $P_1$, which is also one mapping from input space to output space, different one though. This is exactly what a learning machine does: its processing $P$ is adapting. We then have following definition.

\begin{definition}[\bf Universal Learning Machine]
For a learning $\mathcal{M}$, suppose its current processing is $P_0$, and $P_1$ is another processing, if we have one data sequence $T$ (which depends on $P_0$ and $P_1$), so that when we apply $T$ to $\mathcal{M}$, at the end, the processing of $\mathcal{M}$ become $P_1$, we say $\mathcal{M}$ can learn $P_1$ starting from $P_0$. If for any given processing $P_0$ and $P_1$, $\mathcal{M}$ can learn $P_1$ starting from $P_0$, we say $\mathcal{M}$ is an universal learning machine. 
\end{definition}

Simply say, an universal learning machine can learn anything starting from anything. Universal learning machine is desirable. But, clearly, not all learning machine are universal. So, we will discuss what properties can make a learning machine become universal.

In Theorem 1, we gave the relationship of $N$-$M$ IPU and $N$-1 IPU. In order to discuss the relationship of  $N$-$M$ learning machine and $N$-1 learning machine, we need to introduce one property: standing for zero input. We say a learning machine $\mathcal{M}$ with property of standing for zero input, if $\mathcal{M}$ will do nothing for learning, i.e. doing nothing to modify its internal status, when input is zero vector (i.e. $(0, 0, \ldots, 0)$ ) and output side value is empty. Such a property for a learning machine should be very reasonable and very common. After all, zero input means no stimulation from outside, and it is very reasonable to require that learning machine should do nothing for such input.

\begin{theorem}
If we have one $N$-1 universal learning machine $\mathcal{S}$ with property of standing for zero input, we can use $M$ independent $\mathcal{S}$ to construct a $N$-$M$ universal learning machine $\mathcal{M}$.
\end{theorem}
{\bf Proof:} For simplicity and without loss of generality, we only consider the case of $M = 2$. Now, $\mathcal{S}$ is a $N$-1 universal learning machine. As in theorem 1, we can construct a $N$-2 IPU $\mathcal{M}$ by this way:  $\mathcal{M}$ = $(\mathcal{S}_1, \mathcal{S}_2)$. 

$\mathcal{M}$ is sure a $N$-2 learning machine. We only need to show it is universal learning machine. That is to say, for any given processing $P_0$ and $P_1$, there is one data sequence, and driven by the data sequence, $\mathcal{M}$ can learn $P_1$ from $P_0$. 

Actually, we can design a data sequence as following: $T$ is $T_1$ followed by $T_2$, where, $T_1  = ( (D_1, Z_1)$, and $T_2 = (Z_2, D_2) $, where $D_1$ is the data sequence that drives $\mathcal{S}_1$ to learn $P_1$ from $P_0$, $D_2$ is the data sequence that drives $\mathcal{S}_2$ to learn $P_1$ from $P_0$, and  $Z_1, Z_2$ are the zero inputs. Since $\mathcal{S}_1$ and $\mathcal{S}_2$ are universal learning machine, $D_1, D_2$ indeed exist. We know the data sequence $T_1$ followed by $T_2$ indeed is the data sequence we want. 

Of course, the data sequence $T$ ($T_1$ forllowed by $T_2$) is far from optimal, and not desired in practice. But, here we just show the existence of such data sequence.
 
From theorem 1 and 2, we can see that without loss of generality, in many cases, we can focus on $N$-1 learning machine. From now on, we will mostly discuss $N$-1 learning machine. 
\bigskip

{\bf Different Level of Learning } \\
Learning machine modifies its processing by data sequence. Obviously, there is some mechanism inside learning machine to do the learning. More specifically, this learning mechanism would catch information embedded inside data sequence, and use the information to modify its processing. But, we need to be very careful to distinguish 2 things: 1) the learning mechanism only modify the processing, and the learning mechanism itself is not modified; 2) the learning mechanism itself is also modified. But, how to describe these things well?

If $\mathcal{M}$ is an universal learning machine, so, for any giving 2 processing $P_0$ and $P_1$,  we have one data sequence $\mathcal{T}$ so that, starting from $P_0$, and by applying $\mathcal{T}$ to  $\mathcal{M}$, its processing becomes $P_1$. This is clear. But, consider this, somehow, we apply some other data sequence so that the processing becomes $P_0$ again. Since $\mathcal{M}$ is universal, this is allowed. But, we ask, what about if we apply data sequence $\mathcal{T}$ again? what would happen? Do we still have processing becomes $P_1$? There is no guarantee for this. Actually, for many learning machine, this is not the case. However, if this is true, it indicate this: learning mechanism does not change as the processing is changing. This would be one important property. We use next definition to capture this property.

\begin{definition}[\bf Level 1 Learning Machine]
$\mathcal{M}$ is an universal learning machine, for any giving pair of processing $P_0$ and $P_1$, by definition, there is at least one data sequence $T$, so that, starting from $P_0$, and by applying $T$ to  $\mathcal{M}$, processing becomes $P_1$. If the teaching sequence $T$ will only depends on $P_0$ and $P_1$, and dose not depend on any history of processing of $\mathcal{M}$, we call $\mathcal{M}$ as one level 1 universal learning machine.
\end{definition}
Note, following this line of thoughts, we also can define level 0 learning machine, which is an IPU that its processing could not be changed. And, we also can define level 2 learning machine, which is a learning machine that its processing could change, and its learning mechanism could change as well, but its learning mechanism of learning mechanism could not be changed. We can actually follow this line, to define level $J$ learning machine, $J = 0,1, 2, \ldots$. But, we do not discuss in this direction. We will mostly consider level 1 learning machine. 
\bigskip

{\bf Some Examples}\\
{\bf Example 2.1 [\bf Perceptron]}
Perhaps, the simplest learning machine is the perceptron. Perceptron $\mathcal{P}$ is a 2-1 IPU, and it is a learning machine. However, it is not universal. As well known, $\mathcal{P}$ does not have AND gate and XOR gate. That is to say, no matter what, $\mathcal{P}$ could not learn these 2 processing . 
\medskip

{\bf Example 2.2 [\bf RBM is learning machine]} 
See \cite{hinton} for RBM. $N$-1 RBM is one $N$-1 IPU. It is a learning machine as well. There could be many ways to make it learn. The most common way is the so-called Gibbs sampling methods. We can see this clearly: Gibbs sampling is a simple set of rules, and the processing is modified as data is fed into. However, as we can see in Appendix, $N$-1 RBM is not universal.

Put $M$ independent $N$-1 RBM together by the way in theorem 1, we get a $N$-$M$ RBM. So, $N$-$M$ RBM is one learning machine, but it is not universal. 
\medskip

{\bf Example 2.3 [\bf Deep learning might be a learning machine]}
Deep learning normally is a stack of RBM, see \cite{hinton}. It is often formed in this way: first use data to train RBM at each layer, then stack different layers together, then use data to do further training. By the restricted sense, the whole deep learning action is not mechanical learning, since it involves a lot of human intervention. But, if we just see the stage after different layers stacked together, and exclude any further human intervention, it is a mechanical learning. So, in this sense, deep learning is a learning machine.
\medskip

{\bf Example 2.4 [\bf Deep learning might not be a learning machine]}
But, these days, deep learning is much more than stacking RBM together then training without human intervention. There are a lot of pruning, change structure, adjusting done by human. Such learning is surely not mechanical learning. However, many properties can still be studied by point of view of mechanical learning. 
\medskip

Generally, we can say, for software to do learning, it often needs people to establish its very complicated structure and initial parameters. This establishment is not simple and fixed. But, once software is established, and is running without human intervention, such software is learning machine.

\section{Pattern, Examples, Objective and Subjective}\label{table}
Incoming data drive learning. But, IPU and learning machine do not treat data bit-wise. They treat data as patterns. So, patterns are very important to learning machine. Everything of a learning machine is around patterns. Yet, pattern is also quite confusing. We can actually view pattern from different angles and get quite different results. We can view patterns objectively, i.e. totally independent from learning machine and learning, and we can view patterns subjectively, i.e. quite dependent on learning machine and its view on pattern. It is very important we clarify the concept here.
\bigskip

{\bf Examples of Patterns}
Before going to more rigorous discussions, we here discuss some examples of patterns, which could help us to clean thoughts. The simplest patterns are 2-dim patterns. 

{\bf Example 3.1 [\bf All 2-dim Base Patterns]}
2-dim patterns is so simple that we can list all of base patterns explicitly below:
\begin{center}
$PS^0_2$ = $\{ (0,0), (0,1), (1,0), (1,1) \}$ 
\end{center}

All base patterns are here: totally 4 base patterns. For example, (0, 1) is a base pattern. But, besides base patterns, there are more patterns. How about this statement: "incoming pattern is (0,0) or (0,1)"? Very clearly, what this statement describes is not in $PS^0_2$. However, equally clearly, this statement is valid, and specifies an incoming pattern. We have solid reason to believe that the statement represents a new pattern that is not in base pattern space. So, the patterns should be able to include "combination of patterns". We can introduce one way to express this:
\begin{center} 
$p$ = $(0,0) \sqcup (0, 1)$ = \{ one pattern that either (0,0) or (0,1) appears \} 
\end{center} 

In above equation, the symbol $\sqcup$ is called OR (see the similar usage of symbol in \cite{dlog}). The combination operator $\sqcup$ would make a new pattern out of 2 base patterns. Clearly, this new pattern is not in base pattern space. Additional important point: we should note that the new pattern $p$ above is independent from learning machine.

{\bf Example 3.2 [\bf 2x2 Black-White Images]}
We can consider a little more complicated base patterns: 2x2 black-white images. See below illustrations.

\begin{center}
\begin{picture}(340,120)(0,0)
\put(100,20){\resizebox{3 cm}{!}{\includegraphics{Fig1.png}}}
\end{picture}

{\bf Fig. 1 One base pattern in base pattern space of 2x2 black-white images} 
\end{center}

Although in the above illustrations, the patterns are in 2-dim form, it is easy to see that all these patterns can be represented well in linear vector form (for example, the base pattern in Fig. 1 is (1, 1, 0, 1)). It is simple enough so that we can list them:
\begin{center}
$PS^0_4$ = $\{ (0,0,0,0), (1,1,0,0), (0,0,1,1), (1,0,1,0), (0,1,0,1), \cdots \}$ 
\end{center} 

One pattern could be shown as the vector or as 2x2 image. For example, (1,0,1,0) is in vector form, the equivalent image is a vertical line. Let's see some example of combination operators. We can view (1,1,0,0) as one horizontal line, and (0,1,0,1) as one vertical line. Consider this statement "one pattern that has this horizontal line and also this vertical line". Clearly, this is one new pattern. We try to capture it as below:
\begin{center}
$p$ = $(1,1,0,0) \sqcap (0,1,0,1)$ = \{ one pattern that both (1,1,0,0) and (0,1,0,1) appears together \} 
\end{center}  
The symbol $\sqcap$ is called AND (see the similar usage of symbol in \cite{dlog}). But, what is the new pattern $p$? First impression that it is the base patter: $(1,1,0,1)$ (see it in Fig. 1). It is. This is a new base  pattern out from 2 base patterns. How come? Yet, it could be even more complicated. We will address this later.

Now, we should note that the new pattern $p$ above is surely dependent on learning machine and how it views patterns. Without learning machine and how it views patterns, we could not even talk about "appears together".   

We will see another example of pattern but not base pattern. $(1,1,0,0)$ is a base pattern. How about this statement: "one pattern that (1,1,0,0) not appears"? This is one new pattern as well. We would have:
\begin{center}
$p$ = $\neg (1,1,0,0)$ = \{ one pattern that (1,1,0,0) not appears  \} 
\end{center}  
The symbol $\neg$ is called NOT (see the similar usage of symbol in \cite{dlog}). However, what is the new pattern? Is it a group of base patterns: \{(0,0,1,1), (0,0,0,1), $\cdots$\}? As the last question, this should be addressed later.

Besides the above situations, actually, we can see more interesting things (which could not be seen in $PS^0_2$).

{\bf Example 3.4 [\bf Abstraction and Concretization]}
Let's see this pattern:

\begin{center}
$p_h$ = \{ common feature of (1,1,0,0) and (0,0,1,1) \} 
\end{center}  
Clearly, this common feature is not in $PS^0_4$. But, this common feature is one very important pattern: it represents horizontal line. Actually, we can say this pattern $p_h$ is horizontal line. Similarly, we have:  
\begin{center}
$p_v$ = \{ common feature of (1,0,1,0) and (0,1,0,1) \} 
\end{center}  

This time, $p_v$ is vertical line. Further, we can see:
\begin{center}
$p_l$ = \{ common feature of $p_h$ and $p_v$ \}
\end{center}

This time, $p_l$ is line, vertical or horizontal. From the examples above, we can see clearly that abstracting a common feature out from a group of patterns is one very important operation. Without it, we simply could not see some very crucial patterns (such as line). Thus, we need to develop symbols for such operations. For example:

\begin{center}
$p_l$ = $\alpha (p_h, p_v)$
\end{center}

Here, $\alpha$ is one operation that abstract some common features out from the patterns $p_h$ and $p_v$. Note, $\alpha$ is not one operator, but one operation. That means that for same set of patterns, could have more than one operations, which abstract different features from the set of patterns. As we meet more complicated patterns later, this properties would become very clearer.

Very clearly, the operation $\alpha$ is highly dependent on a learning machine and what the learning machine learned previously. 

Conversely to abstraction operation $\alpha$, we can also have concretization operation $\rho$. See examples below:
\begin{center}
$p = \rho(p_h, (0,0,0,1))$ = \{ one concrete horizontal line related to the pattern (0,0,0,1) \} = (0,0,1,1) 
\end{center}

$\rho$ is one operation that concretize a pattern (which is one abstraction pattern) by related it to some pattern. 
Any concretization of a pattern is a pattern. As above, concretizing a horizontal line would give a real horizontal line. And, since it is related to (0,0,0,1), this horizontal line should be (0,0,1,1). 

Very clearly, the operation $\alpha$ and $\rho$ are highly dependent on a learning machine (such as: what the learning machine learned previously, how it views patterns, etc). 

From above examples, we can see that patterns are much more than base patterns. We can have pattern of patterns (see horizontal lines, vertical lines). We can have pattern of patterns of patterns (see line). We can have operations on the patterns. We have operators of patterns. All results are still patterns. So, patterns are not just one type, it has many types. Or, we can say patterns are typeless. Base patterns are just simplest patterns and fundamental building blocks.

{\bf Example 3.3 [\bf 4x4 Black-White Images]}
We now consider even more complicated patterns: 4x4 black-white images. See below illustrations.

\begin{center}
\begin{picture}(340,120)(0,0)
\put(100, 20){\resizebox{3 cm}{!}{\includegraphics{Fig2.png}}}
\end{picture}

{\bf Fig. 2  A base pattern in base pattern space of 4x4 black-white images} 
\end{center}

The binary vector space has $2^{16}$ elements. This is a large number. While in theory we can still list all base patterns, it would be very hard.  
\begin{center}
$PS^0_{16}$ = $\{ \cdots, (1,1,1,1,0,0,0,0,0,0,0,0,0,0,0,0,0,0,0,0), \cdots \}$ 
\end{center} 

Since there is a larger dimension, more phenomenon would appears. We can see  some of them here. Clearly, the binary vector shown in the above equation is one horizontal line. So, we can still have:
\begin{center}
$p_h$ = $\alpha$(first 2 horizontal lines)
\end{center}  

Clearly, this pattern $p_h$ is not in $PS^0_{16}$. But, it represents first 2 horizontal lines. Can this pattern $p_h$, which abstracts first 2 horizontal line, represent all horizontal 4 lines? This is one very important question. At this moment, we can not answer it. 

Similarly, we have:  
\begin{center}
$p_v$ = $\alpha$(all vertical lines) 
\end{center}  

And, we ca have: 
\begin{center}
$p_l$ = $\alpha (p_h, p_v)$
\end{center}

But, again, since we are dealing more complicated patterns space now, we can see something that Example 2 could not show. How about: 
\begin{center}
$p_c$ = \{ a point at coordination (3,3) \},  $p_b$ = \{ a point at coordination (0,0) \}
\end{center} 

\begin{center}
$p_0$ = $\rho(p_v, p_b)$
\end{center} 
This is concretization of vertical line related to point (0,0). 

And, more:
\begin{center}
$p$ = $p_0 \sqcap p_c$
\end{center} 
This is one pattern with one vertical line and a point at (3,3). The pattern $p$ is AND of 2 different types of patterns. This is one example that we have to make all operations and operators on patterns typeless.

Let's try to put the above equations together, we then have:
\begin{center}
$p$ = $\rho(\alpha$(all vertical lines), \{ a point at coordination (0,0) \}) $\sqcap$ \{ a point at coordination (3,3) \}
\end{center}

Might be easier to just state: a vertical line pass through (0,0) and a point at (3,3). But, as we can see, the above equation describe the pattern much more precisely and mechanically (i.e. to avoid to use language, either natural language or programming language, just use our simple and mechanical terms: $\alpha$, $\rho$, $\sqcup$, $\sqcap$, $\neg$).

We examined some simple examples above. Though simple, they are very revealing. From these examples, we can see some important properties of patterns. First, patterns are more than base patterns, much more. Second, some patterns together could generate new pattern. There are many ways to generate new patterns, such as OR, AND, NOT, abstraction, concretization, and more. Third, very crucially, we realize that some patterns are independent from learning machine, while some depend on learning machine heavily. In other words, for a learning machine, some patterns are objective, while some are subjective. 
\bigskip

{\bf Pattern, Objectively } \\
First, we want to discuss pattern that is objective to learning machine. Base pattern is the foundation for all patterns. We defined it before. But, we repeat it again here for easy to cite.

\begin{definition}[\bf Base Pattern Space]
$N$-dim base pattern space, denote as $PS^0_N$,  is a $N$-dim binary vector space, i.e.  
\end{definition}
\begin{center}
$PS^0_N$ = $\{ all \  p = (p_1, p_2, \ldots, p_N) | \  p_k = 0 \ or \ 1 \}$ 
\end{center}

Each element of $p \in PS^0_N$ is a base pattern. There are totally $2^N$ patterns in $PS^0_N$. When $N$ is not very small, $PS^0_N$ is a huge set. Actually, this hugeness is the source of richness of world and fundamental reason of difficulty of learning.

Base pattern space is just the starting point of our discussion. From above examples, we know that many patterns are not base pattern. But, if a pattern is not base pattern, what is it? We can see in this angle: no matter what a pattern is, what is presented to input space of a learning machine is a base pattern. So, naturally, we have definition below.

\begin{definition}[\bf Pattern as Set of Base Patterns]
A $N$-dim pattern $p$ is a set of base patterns: 
\end{definition}
\begin{center}
$p$ = \{ $b_1, b_2, \cdots | b_i \in  PS^0_N$ \}
\end{center}
We can denote this set as $p_b$, and call is as the base set of $p$ (b stands for base). While we use $p$ as the notation of a pattern, we understand it is a set of base patterns. If we want to emphasis it is a set of base patterns, we use notation $p_b$. We also can write $p = p_b$. Any base pattern in base set is called a base face of $p$ (or just simply face). For example, in above, $b_2$ is one face of $p$. Specially, any base pattern $b$ is one pattern, and it is the (only) base face of itself. 

According to this definition, a pattern is independent from learning machine, which is just a group of base patterns, no matter what a learning machine is. If we want to view pattern objectively, the only way is to define a pattern as a group of base patterns. So, objectively, a pattern is a set of base patterns. 

What objective operators on objective patterns are? Since patterns are set of base patterns, naturally we first examine basic set operations: union, intersection, and complement.

\begin{definition}[\bf Operator OR (set union)]
Based on any 2 patterns $p_1$ and $p_2$ , we have a new pattern $p$: 
\end{definition}
\begin{center}
$p$ = $p_1$ OR $p_2$ = $p_{1_b} \cup p_{2_b}$ 
\end{center}

Here, $\cup$ is the set union. That is to say, this new pattern is such a pattern whose base set is the union set of base sets of 2 old patterns. Or, we can say, $p$ is such a pattern whose face is either a face of $p_1$ or a face of $p_2$.

\begin{definition}[\bf Operator AND (set intersection)]
For any 2 patterns $p_1$ and $p_2$ , we define a new pattern: 
\end{definition}
\begin{center}
$p$ = $p_1$ AND $p_2$ = $p_{1_b} \cap p_{2_b}$ 
\end{center}
 
Here, $\cap$ is the set intersection. Or we can say, $p$ is such a pattern that its face is both face of $p_1$ and $p_2$. In this sense, we say, $p$ is both $p_1$ and $p_2$.

\begin{definition}[\bf Operator NOT (set complement) ]
For any patterns $p$ , we define a new pattern: 
\end{definition}
\begin{center}
$q$ = NOT $p$ = $p_{b} ^c$ = $\{ b \in PS^0_N | b \not \in p_{b} \}$  
\end{center}

Here, $A^c$ is complement set of $A$. That is to say, $q$ is such a pattern that its face is not a face of $p$.  

Very clearly, the above 3 operators do not depend on learning machine. So, they are all objective. Consequently, if we apply these 3 operators consecutively any times, we still generate a new pattern that is objective. 
\bigskip

{\bf Pattern, Subjectively} \\
Now we turn attention to subjective pattern, i.e. pattern to be viewed from a particular learning machine.

We need to go back for a while and consider basic. When we say there is an incoming pattern $p$ to a learning machine, what do we mean? If we see this objectively, the meaning is clear: at input space, a binary vector is presented, which is a face of the incoming pattern $p$. This does not depend on learning machine at all. And, this is very clear and no unambiguity.  

However, as our examples demonstrated, we have to consider pattern subjectively. We need to go slowly since there are a lot of confusing here. We have to consider something that is not valid at all objectively. 

{\bf Pattern, 1-significant or 0-significant }\\
First, when we discuss patterns subjectively, we need to know: Is 1 significant? or 0 significant, or both are equally significant?

Does this sound wrong? By definition, a base pattern is a binary vector, so, of course, both 0 and 1 would be equally significant. Why consider 1-significant, or 0-significant? Let's consider one simple example. For 4-dim pattern, $p_1 = (1, 1, 0, 0)$ is one base pattern, and could be viewed as one horizontal line (see example 2 and Fig. 2). $p_2 = (0, 1, 0, 1)$ is also one base pattern, and could be viewed as one vertical line. When we talk about $p_1$ and $p_2$ appears together (or happen together), do we mean this pattern: (1, 1, 0, 1), or (0, 1, 0, 0)? Former one is 1-significant, and latter is 0-significant. So, if we want to use the term such as "2 pattern happen together", it is necessary to distinguish 1-significant and 0-significant. 

So, to distinguish 1-significant pattern or 0-significant pattern indeed makes sense, and is necessary. When we consider a pattern as 1-significant, we often look at its 1 components, not pay much attention to its 0 components, just as we did in the example: "(1, 1, 0, 1) equals (1, 1, 0, 0) and (0, 1, 0, 1) appear together". Contrast, we do not think:  "(0, 1, 0, 0) equals (1, 1, 0, 0) and (0, 1, 0, 1) appear together", since we do not consider 0-significant. 

Perhaps, 1-significant is actually already in our sub-conscious. Just see which sentence is more appealing to us: "(1, 1, 0, 1) equals (1, 1, 0, 0) and (0, 1, 0, 1) appear together", or "(0, 1, 0, 0) equals (1, 1, 0, 0) and (0, 1, 0, 1) appear together". 

Additional to the above consideration, most patterns that people consider for many applications are sparse pattern, i.e. only a few bits in the pattern are 1, most are zero. For sparse patterns, 1-significant is very natural choice. In fact, in sparse pattern, 1-significant is very natural. Just see this example:
\begin{center}
(1, 1, 0, 1, 0, 0, 0, 0, 0, 0, 0, 1) =  \\ (1, 1, 0, 0, 0, 0, 0, 0, 0, 0, 0, 0)  and (0, 1, 0, 1, 0, 0, 0, 0, 0, 0, 0, 1) appear together.
\end{center}
We would accept this statement easily. 
From now on, unless we state explicitly, we will use 1-significant. \\

{\bf Patterns and Learning Machine} \\
From the examples, we know that one pattern $p$ could be perceived very differently by different learning machine. This make us consider this question carefully: from the view of learning machine, what really is a patterns? We have not really addressed this crucial question yet, we just let our intuition play at the background. In Example 2, when we talk about $\sqcap$ operator, and give an equation $p$ = $(1,1,0,0) \sqcap (1,0,1,0)$, we did not really tell what is this pattern $p$. Now we address this more carefully.

Take a look at this: {\it \{ one pattern that both (1,1,0,0) and (1,0,1,0) appears together \} }. In our tuition, this is a right thought. However, if we see things objectively, this is simply wrong: Base patterns (1,1,0,0) and (1,0,1,0) cannot appears together. They are different base patterns. At one time, only one of them can appear. In this sense, "together" cannot happen.

To address this question, we have to going deep to see what a pattern really is. When we talk about base pattern, i.e. binary vector in $PS^0_N$, there is no unambiguity. Everything is very clear. However, just base pattern is not good enough. With only base patterns, we simply cannot handle most things that we want to work with. 

At this point, we should be able to realize that pattern is not only associated with what is presented at input space (surely base pattern), but also associated with how a learning machine perceives incoming pattern. For example, when base pattern (1,1,1,0) is at input, the learning machine could perceive it is just one base pattern, but also could perceive it as two base patterns (1,1,0,0) and (1,0,1,0) appear together, or could perceive much more, much more complicated. 

So, naturally, a question arise: can we define pattern without introducing perception of learning machine? Yes, this can be done. Since no matter what pattern is, when a pattern is sent to learning machine, it is one base pattern at input space. In this way, we can surely define a pattern to be a set of base patterns. So, no matter what learning machine is and how it perceives, pattern is a set of base patterns. This is just objective pattern. For example, we can forcefully define  {\it \{ one pattern that both (1,1,0,0) and (1,0,1,0) appears together \} } as the set of base patter  \{ (1,1,1,0) \}. This is what we did in above section. 

Seems this way resolves unambiguity. However, as all examples indicated, objective way cannot go far and we need to understand patterns subjectively. Pattern cannot be separated from how a learning machine perceives. Pattern defined as a set of base patterns is precise, but how a learning machine perceives patterns is much more important. Without learning machine perceiving, actually, no matter how precise a pattern is, it is not much useful.

Here, it is worth to stop and review our thoughts here. The major point is: learning machine plays an active role, and it must have its own way to see its outside world. More precisely, a learning machine must have the ability to tell what is outside of itself, and what is inside of itself, and what is its view to outside. With or without such ability is very critical. Only with this ability, the learning machine can go further and our later discussions can be conducted. It is very important we realize this. Without such ability, a learning machine is reduced to an ordinary computer program that is very hard to do learning. From now on, our learning machine will have such ability and we will make the ability more clearer. So, patterns would be mainly subjective to a learning machine. 

Thus, we have to address this critical issue: how a learning machine perceives pattern? And we need to see this by considering relationship among patterns. We need to think these issues as well: 1) how to form new pattern from old pattern? 2) how to associate new pattern with prior learned patterns? 3) how to organize learned patterns? 4) how to re-organized learned patterns? In order to do these, we have to see how machine perceives. 
\bigskip

{\bf How Learning Machine Perceives Patterns} \\
How a learning machine perceives pattern is closely related to how it processes information. So we go back to IPU for a while. Consider a $N$-1 IPU $\mathcal{M}$, suppose its processing is $\mathcal{P}$. We define of black set:

\begin{definition}[\bf Black set of $N$-1 IPU]
For a $N$-1 IPU $\mathcal{M}$, if its processing is $\mathcal{P}$, the black set of $\mathcal{M}$ is:
\begin{center}
$B$ = $\{ b \in PS^0_N | \mathcal{P}(b) = 1 \}$ 
\end{center}
\end{definition}
Equivalently, we also call $B$ as the black set of processing $\mathcal{P}$.

For IPU $\mathcal{M}$, suppose $B$ is its black set, this means: if we put one base pattern $b \in B$ to input space, $\mathcal{M}$ will process it to 1, if $b \notin B$, to 0. This reveals one important fact to us: inside $\mathcal{M}$, there must exist a bit $pb$ with such property: if input $b \in B$, $pb = 1$, if $b \notin B$, $pb = 0$. 

We do not know exactly what inside $\mathcal{M}$ is, and we do not know how exactly the processing is done. However, we do know such a bit $pb$ must be there. We do not know where this bit $pb$ is and exists in what form, but we know it exists. Otherwise, how can $\mathcal{M}$ be able to distinguish input from $B$ or not from $B$? Such bit $pb$ reflects how $\mathcal{M}$ process input to output. We can imagine that $\mathcal{M}$ could have more such bits. So, we have definition.

\begin{definition}[\bf Processing Bits]
For IPU $\mathcal{M}$, if it has some internal bit $pb$ has such properties: there is a set $B \subset PS^0_N$ so that for any $b \in B$, $pb = 1$ (light), for any $b \not\in B$, $pb = 0$ (dark), we call bit $pb$ as one processing bit. If $\mathcal{M}$ has more than one such bit, say, $pb_j, j=1, \ldots, L$ are all processing bits, we call such set as set of processing bits of  $\mathcal{M}$, or just simply, processing bits.
\end{definition}

\begin{theorem}
For a IPU $\mathcal{M}$, set of processing bits $\{pb_j \ |\ j=1, \ldots, L\}$ exists and is not empty.
\end{theorem}
{\bf Proof: } We will exclude 2 extreme cases, i.e. $\mathcal{M}$ maps all input to 0 vector $(0, 0, \ldots, 0)$, and $\mathcal{M}$ maps all input to 1 vector $(1, 1, \ldots, 1)$. After excluding the 2 extreme cases, we can say, black set $B$ of $\mathcal{M}$ is a proper subset of $PS^0_N$, so does $B^c$. Thus, as we argued above, there must exist a bit $pb$ inside $\mathcal{M}$ with such property: for $b \in B$, $pb = 1$, for $b \notin B$, $pb = 0$. So, set of processing bits indeed exists and not empty.
 
In proof, we show that set of processing bits are not empty, at least there is bit in it. Such case indeed exists. There are IPU whose set of processing bits has only one elememt. But in most cases, set of processing bits has more than one element. In fact, $L$, the number of processing bits, can reflect the complexity of IPU. Processing bits reflects how processing of IPU is conducted.

Since a learning machine is also a IPU, it has processing bits as well. But, as we discussed before, how a learning machine perceives pattern is closely related to how it process input. So, for learning machine, we will call these bits as perception bits, instead of processing bits. When one base pattern is put into input, each perception bit will take its value. All these values together, we have perception values. Perception values reflects how a learning machine perceives this particular base pattern. If a learning machine is learning, its perception bits could chang, the number perception bits could increase or decrease, its behavior could change. Even the array of perception bits might not change, the behavior could change. 

Armed with perception bits, we can describe how $\mathcal{M}$ perceives pattern. When a base pattern $b$ is put into input space, perception bits act, some are light and some are dark. These bits reflect how $b$ is perceived, i.e. the perception bits $\{pb_j \ |\ j=1, \ldots, L\}$ are taking values, we have a binary vector $pv = (pv_1, pv_2, \ldots, pv_L)$, where $pv_j$ is value (1 or 0) of $pb_j$ takes. We call them as perception values. Note, the perception values depends on a particular base pattern. The perception values tells how $\mathcal{M}$ perceives a base pattern $b$. 

If $b$ and $b'$ are 2 different base patterns, i.e. they are different bit-wise, but they have same perception values, we know that these 2 base patterns are perceived as same by $\mathcal{M}$, since $\mathcal{M}$ has no way to tell any difference between $b$ and $b'$. If 2 base patterns are possibly perceived different by $\mathcal{M}$, their perception values must be different (at least one perception bit must behaves differently).

However, reverse is not true. It is possible that 2 base patterns $b$ and $b'$ have different perception values, but $\mathcal{M}$ still could perceives $b$ and $b'$ as same subjectively. That is to say, $\mathcal{M}$ can perceives 2 different base patterns as same even their perception values are different. So we have definition below.

\begin{definition}[\bf Base patterns perceived same by a learning machine subjectively]
Suppose $\mathcal{M}$ is a learning machine, $\{pb_j \ |\ j = 1, \ldots, L\}$ are perception bits, if for 2 base patterns $b_1$ and $b_2$, their perception values are $(pv^1_1, pv^1_2, \ldots, pv^1_L)$ and $(pv^2_1, pv^2_2, \ldots, pv^2_L)$, and at least for one $k$, $pv^1_k = pv^2_k = 1$, we say, at perception bit $pb_k$, $\mathcal{M}$ could subjectively perceives $b_1$ and $b_2$ as same. 
\end{definition}
That is to say, for 2 base patterns, if at any perception bit, their perception values are different, learning machine is not possible to perceive them same. But, if at least at one perception bit, their perception values are both 1, $\mathcal{M}$ could possibly perceive them as same subjectively. Of course, $\mathcal{M}$ could also perceive them as different subjectively. Note, perception value should be 1, not 0. This is related to 1-significant.

\begin{definition}[\bf Pattern perceived by a learning machine subjectively]
Suppose $\mathcal{M}$ is a learning machine, $\{pb_j \ |\ j = 1, \ldots, L\}$ are perception bits. And suppose $p$ is a group of base pattern, and  at perception bit $pb_k, 1\le k \le L$, perception value for all base patterns in $p$ equals 1, then $\mathcal{M}$ could perceive all base patterns of $p$ as same, and if so, we say $\mathcal{M}$ perceives $p$ as one pattern subjectively at $pb_k$, and $p$ forms a subjective pattern. 
\end{definition}
Note, in definition, it only needs that all base patterns in $p$ behaves same at one perception bit. This is minimal requirement. Of course, this requirement could be increased. For example, to require at all perception bits behaving same. But, all requirements are subjective.

Here we put down the major points about subjective patterns and how a learning machine to perceive them.
\begin{enumerate}
\item  There are perception bits in a learning machine (only exclude 2 extreme cases). Any system that satisfies the definition of Iearning machine must have perception bits. How perception bits are formed and how exactly perception bits are realized inside a learning machine could be different greatly. But we emphasis that perception bits indeed exist. 

\item These bits are very crucial for a learning machine. They reflect how learning machine perceive and process patterns. When a base pattern is put into input space of learning machine, then perception bits act and the learning machine uses these values to perceive pattern subjectively, and process pattern accordingly. 

\item For learning machine, its perception bits are changing with learning. However, even the number of perception bits are not changing, the behavior of perception bits could change (so does the perception of learning machine). 

\item Armed by perception bits, we can well understand subjective pattern. If 2 base patterns behave same at one perception bit, then, the 2 base patterns can be perceived as same at this perception bit subjectively. This can be extended to more than 2 base patterns. For a group of base patterns $p$, and if all base patterns behave same at one perception bit, then $p$ can be perceived as same at this perception bit subjectively. This is the way to connect objective and subjective.
\end{enumerate}

To consider pattern objectively, only need to involve set operation, no need to do any modification on learning machine itself. But, to consider pattern subjectively, set operation could be used. But more importantly, perception bits are needed. And, quite often, to modify perception bits is necessary. For subjective operator of subjective patterns, we need to base our discussion on perception bits.
\bigskip

{\bf Pattern, Subjective Operators} \\
Just as operators for objective patterns, it is naturally to consider subjective operators for subjective patterns. There are 3 very basic operators: NOT, OR, AND. First, consider NOT.

\begin{definition}[\bf Operator NOT for a Pattern, Subjectively]
Suppose $\mathcal{M}$ is a learning machine, $\{pb_j \ |\ j = 1, \ldots, L\}$ are perception bits. For a subjective pattern $p$ perceived at $pb_k$ by $\mathcal{M}$, $q$ is another pattern perceived at $pb_k$ by $\mathcal{M}$ in this way: $q$ are all such base patterns that is perceived by $\mathcal{M}$, and at $pb_k$, the perception value is 0. 
\end{definition}
We can denote this pattern $q$ as $q$ = NOT $p$ or $q = \neg p$. This notation $\neg$ is following \cite{mkrot}. We can also say, pattern $q$ is a pattern that $p$ does not appear. 

Note, this operation NOT is subjective. $q$ consists of base patterns that are perceived by $\mathcal{M}$. So, this is quite different than the objective operation NOT (set complement). Another important point is: in order to do this operator, no need to modify perception bits of $\mathcal{M}$, only perception value is different. 

Now we turn attention to another operator OR. Consider that we have a subjective pattern $p_1$, and the perception values of $p_1$ are $pv^1_1, \ldots$, and subjective pattern $p_2$, and the perception values of $p_2$ are  $pv^2_1, \ldots$. Since $p_1$ and $p_2$ are different pattern, their perception values must be different at some bits. Now, we want to put them together to form a new pattern, i.e. $p = p_1 OR p_2$, which measn either $p_1$ or $p_2$. This action of course changes the perception of $\mathcal{M}$ and must change the perception. If the perception is not changed, there is no way to have OR. So, when we introduce the OR operator, we in fact change $\mathcal{M}$. This is what subjective really means: learning machine changes its perception so that $p_1$ and $p_2$ are treated same, though $p_1$ and $p_2$ indeed have difference, and the difference is ignored.

\begin{definition}[\bf Operator OR for 2 Patterns, Subjectively]
Suppose $\mathcal{M}$ is a learning machine, $\{pb_j \ |\ j = 1, \ldots, L\}$ are perception bits. For any 2 subjective patterns $p_1$ and $p_2$, $p_1$ perceived at $pb_{k_1}$ by $\mathcal{M}$, and $p_2$ perceived at $pb_{k_2}$ by $\mathcal{M}$, $p$ is another subjective pattern, and perceived by $\mathcal{M}$ in this way: first $\mathcal{M}$ will modify its perception bits if necessary, then $\mathcal{M}$ perceive any base patterns from either $p_1$ or $p_2$ at another perception bit $pb_l$ same. That is to say, if $pb_l$ does not exist, $\mathcal{M}$ will generate this perception bit first.
\end{definition} 
We can also say, new pattern $p$ is either $p_1$ or $p_2$ appears. We can denote this new pattern as $p = p_1 OR p_2$ = $p_1 +_s p_2$. This notation $\neg$ is following \cite{valiant}.

Note, if we want to do operation OR, we might need to modify perception bits of $\mathcal{M}$. This is often done by adding a new perception bit. This is totally different from the objective OR (or set union). On surface, $p_1 + p_2$ indeed is a union (set union) of $p_1$ and $p_2$. But, without modification of perception bits, there is no way to do this union.

Then consider subjective operator AND. This operator is crucially important. Actually, we spent a lot of time to argue about this operator, i.e. {\it appears together}.

\begin{definition}[\bf AND Operator for 2 Base Patterns, Subjectively]
Suppose $\mathcal{M}$ is a learning machine, $\{pb_j \ |\ j = 1, \ldots, L\}$ are perception bits. If  $p_1$ is one subjective patterns perceived at $pb_{k_1}$, $p_2$  is one subjective patterns perceived at $pb_{k_2}$, then, all base patterns that $\mathcal{M}$ perceives at both $pb_{k_1}$ and $pb_{k-2}$ at the same time will form another subjective pattern $p$, and $p$ is perceived by $\mathcal{M}$ at $pb_l$. That is to say, if $pb_l$ does not exist, $\mathcal{M}$ will generate this perception bit first.
\end{definition}
We can also say, new pattern $p$ is both $p_1$ and $p_2$ appear together. We can denote this pattern $p$ as $p = b_1 \ AND \ b_2$ = $b_1 \cdot b_2$. This notation $\cdot$ is following \cite{valiant}.

Note, if we want to do AND operator, we have to modify perception bits of $\mathcal{M}$. This is totally different from the objective AND (or set intersection). 
\bigskip

{\bf X-Form} \\
We have setup 3 subjective operators for subjective patterns. If applying the 3 operators consecutively, we will have one algebraic expression. Of course, in order this algebraic expression makes sense, learning machine needs to modify its perception bits. But, we want to know what we can construct from such algebraic expressions? First, we see some examples.

{\bf Example 3.4 [\bf One Simple X-form ]}
Suppose $b_1, b_2, b_3$ are 3 different base patterns. Then,
$$
E(b_1, b_2, b_3) = b_1 + (b_ 2 \cdot b_3)
$$
is one subjective pattern. We can say, this pattern is: either $b_1$ or $b_2$ and $b_3$ happen together. However, the expression has more aspects. Since $E$ is one algebraic expression, we can substitute base patterns into it, and get one value. This is actually what algebraic expression for. That is to say, $E(b) = b_1 + b_3$ is one mapping on $PS_N^0$ to \{0, 1\}, and it behaves like this: for any $b \in PS_N^0$, if $b = b_1$ or $b \in b_2 \cdot b_3$, $E(b) = 1$, otherwise $E(b) = 0$. This matches our intuition well.

{\bf Example 3.5 [\bf More X-forms ]}
If $g = {b_1, b_2, \ldots, b_K}$ is a group of base patterns, and we have some algebraic expressions, we get more subjective patterns based on $g$. See some example here: 
$$
b_1 + b_3, \ b_2 + (\neg b_1 + b_2 \cdot b_5), \ (b_2 + (b_1 \cdot b_2)) \cdot (b_3 + b_4), \ \ldots
$$
are subjective pattern. But, also these expressions can be used to define a mapping on  $PS_N^0$ to \{0, 1\}, just like above.

{\bf Example 3.6 [\bf Prohibition ]}
If $e_1$ and $e_2$ are expressions, we want to find an expression for this situation: $e_2$ prohibits $e_1$, i.e. if $e_2$ is light, output has to be dark, otherwise, output equals $e_1$. This expression is:
$$
\neg (e_1 \cdot e_2) + (e_1  \cdot \neg e_2)
$$
are subjective pattern.

Above, each expression has 2 faces: first, it is one algebraic expression, second, it is one subjective pattern perceived by $\mathcal{M}$. In order to make sense for these expressions, $\mathcal{M}$ has to modify its perception bits accordingly. This is crucial. Thus, we have following definition.

\begin{definition}[\bf X-Form for patterns]
If $E$ is one algebraic expression of 3 subjective operators, and $g = \{b_1, b_2, \ldots, b_K\}$ is a group of base patterns, then we call the expression $E(g) = E(b_1, b_2, \ldots, b_K)$ as an X-form upon $g$, or simply X-form. We note, in order to have this expression make sense, quite often, learning machine $\mathcal{M}$ needs to modify its perception bits accordingly. And, if this expression make sense, we then have a subjective pattern $p = E(g) =  E(b_1, b_2, \ldots, b_K)$.
\end{definition}
The name X-form is chosen for reason: these expressions are forms, and we do not know them well, and X means unknown. In \cite{valiant}, there are similar form called conjunction normal form (CNF). Though, our expression are quite different than CNF of Valiant. CNF of Valiant is basically objective, while X-forms are subjective. 

One important aspect of X-form is: it is one algebraic expression, so, we can substitute variable into and calculate to get output value, 0 or 1. See above examples. In this sense, one X-form would be a mapping on $PS_N^0$ to \{0, 1\}. The calculation of this expression is actually same as learning machine doing processing inside itself. This is one wonderful property. This is exactly the reason why we introduce the construction of X-form. In this way, one X-form can be thought as one processing. Thus, we can also think one X-form has a black set, which is exactly equals the subjective pattern of this X-form.   

In order to connect objective patterns, subject patterns, and X-form, we have following theorem.

\begin{theorem}
Suppose $\mathcal{M}$ is a $N$-1 learning machine. For any objective pattern $p_o$ (i.e. a set of base patterns in $PS_N^0$),  we can find some algebraic expression $E$ upon some group of base patterns $g = {b_1, b_2, \ldots, b_K}$ so that $p_o = E(g)$. If so, we say X-form $E(g)$ express $p_o$. In most cases, there are many X-form to express $p_o$. However, among those X-forms, we can find at least one so that it base upon no more than $N$ base patterns, i.e. in $g = \{b_1, b_2, \ldots, b_K\}$ $K \le N$. 
\end{theorem}
{\bf Proof:} Suppose $p_o$ is one objective pattern. It is easy to see there is one algebraic expression $E$ can express $p_o$. Since $p_o$ is a set of base patterns, surely we can write $p_o$ as:
$$
p_o = \{b_1, b_2, \ldots, b_L \}, \ b_i \in PS_N^0
$$ 
where each $b_i$ is a base pattern. The algebraic expression
\[
E_1(b_1, b_2, \ldots, b_L) = b_1 + b_2 + \ldots + b_L   \label{eq:ex1} \tag{1}
\]
can express $p_o$, since we can easily see $p_o = E_1(b_1, b_2, \ldots, b_L)$. If $L$ is not bigger than $N$, we already find such a group of base pattern and such an algebraic expression, and proof is done. 

If $L$ is bigger than $N$, we can do further. For one base pattern $b$, we can find some other base patterns $b'_1, b'_2, \ldots, b'_J$, $J \le N$, and to express $b$ in this way:  $b = b'_1 \cdot  b'_2 \cdot \ldots \cdot b'_J$. Such $b'_1, b'_2, \ldots$ sure can be found. For example, if $b = (1, 1, 0, \ldots, 0)$, we can find $b'_1 = (1, 0, \ldots, 0)$ and $b'_2 = (0, 1, 0, \ldots, 0)$, then $b = b'_1 \cdot b'_2$.

For a group of base patterns, we can do same. That is to say, for $b_1, b_2, \ldots, b_L$, we can find There are at most $N$ base patterns $b'_1, b'_2, \ldots, b'_K$, $K \le N$ so that for each of $b_j, j = 1, 2, \ldots, L$, we can  find some $b'_{j_1}, \ldots, b'_{j_{K_j}}, K_j \le K$, and $b = b'_{j_1} \cdot \ldots \cdot b'_{j_{K_j}}$. We know such a group base patterns indeed exists. For example, $b'_1 = (1, 0, \ldots, 0, 0), \ldots, b'_N = (0, 0, \ldots, 0, 1)$ are such a group.

Now, we can continue: 
\begin{multline}
p_o = E_1(b_1, b_2, \ldots, b_L) = b_1 + b_2 + \ldots + b_L =  \\ 
(b'_{1_1} \cdot \ldots \cdot b'_{1_{K_1}}) + (b'_{2_1} \cdot \ldots \cdot b'_{2_{K_2}})  +  \ldots +  (b'_{L_1} \cdot \ldots \cdot b'_{L_{K_L}}) = \\
E(b'_1, b'_2, \ldots, b'_K)  \label{eq:ex2}  \tag{2}
\end{multline}

This algebraic expression $E$ and a group of base patterns $b'_1, b'_2, \ldots, b'_K$, and $K \le N$, are what we are looking for. We should note, expression \eqref{eq:ex1} is "level 1" expression, while \eqref{eq:ex2} is "level 2" expression. We can do for higher level expressions.

Of course, the expression in the proof is just used to do the existence proof. It is not best expression. This expression is very "shallow". We can push the expression to higher level. But, here we do not discuss how to do so.

Theorem 4 tells the relationship of objective pattern and subjective pattern. For any objective pattern $p_o$, we can find a good group of base patterns (size of this group is as small as possible, at worst, not greater than $N$), and a good algebraic expression, to express this objective pattern as one subjective pattern. 

Here is the major point. One objective pattern $p_o$ is a set of base patterns. However, when $p_o$ is perceived by a learning machine, learning machine generates a subjective pattern. The major question is: will the subjective one match with objective one? Theorem 4 confirm that, yes, for any objective pattern $p_o$, we always can find X-form to express $p_o$. 

Naturally, next we would ask, how well such expression is? For "how well", we need some criteria. There could have many such criteria. However, this criteria is very important: use as less as possible base patterns, i.e. in $p_o = E(b) = E(b_1, b_2, \ldots, b_K)$, $K$ is as small as possible. There could have other important properties of X-form. To satisfy these properties, we can get a better X-form.

Of course, next question is how to really find or construct such X-form. That is what we do next. 
\bigskip

{\bf Sub-Form} \\
Several X-form could form a new X-form. And, some part of a X-form is also a X-form. Such part could be quite useful. So, we discuss sub-form here.

\begin{definition}[\bf Sub-Form of a X-form]
Suppose $e$ is a X-form, so, it is one algebraic expression $E$ (of 3 subjective operations) upon a set of base patterns $g = \{b_1, b_2, \ldots, b_K\}$ so that $e = E(g) = E(b_1, b_2, \ldots, b_K)$. A sub-form of $e$ is one algebraic expression $E_s$ upon a subset of $g$, $g_s = \{b_{s_1}, \ldots, b_{s_J}\}$,  $J \le K$, so that $e_s = E_s(g_s) = E_s(b_{s_1}, \ldots, b_{s_J})$, and the objective pattern expressed by $e_s$ is a proper subset of the objective pattern expressed by $e$.
\end{definition}
So, by definition, a sub-form is also a X-form. 

{\bf Example 3.7 [\bf  Sub-Form]}
1. $e = b_1 + b_2$ is one X-form. Both $b_1$ and $b_2$ are sub-form of $e$. \\
2. $e = b_1 + b_2 \cdot b_2$ is one X-form. Both $b_1$ and $b_2 \cdot b_3$ are sub-form of $e$. But, $b_2$ (or $b_3$) is not sub-form of $e$. \\
3. $e = (b_1 + b_2) \cdot (b_1 + b_3)$ is one X-form. We can see that the black set of $e$ is $\{b_1, b_1 \cdot b_2, b_1 \cdot b_3, b_2 \cdot b_3 \}$. So, $b_1$ is sub-form of $e$, but $b_2, b_3$ are not.

One X-form $e$ could have more than one sub-form. Or one X-form has no sub-form. For a sub-form, since it is one X-form, it could have sub-form for itself. So, we can have sub-forms for sub-forms, and go on. It is easy to see, any sub-form of sub-form is still a sub-form. So, X-form could have many sub-form. We denote all sub-forms as $e_i, i = 1, 2, \ldots L$. These sub-form are play important roles. They are actually fabric of processing.

\section{Learning by Teaching}\label{table}
We now turn attentions to learning. We emphasis again that a learning machine is based on patterns, not bitwise, and the purpose of a learning machine is to process patterns and learn how to process. 

Theorem 1 and Theorem 2 tell us, for simplicity and without loss of generality, we can just consider $N$-1 learning machine. For a $N$-1 learning machine, its processing is actually equivalent to its black set. We can also consider an objective pattern $p$, which is a set of base patterns. Thus, $p$ can be thought as black set of one processing, and vise versa. This tells us, for a $N$-1 learning machine, its processing is equivalent to a objective pattern, called its black pattern. Obviously, black set and black pattern are equivalent. We can switch the 2 terms freely. By this understanding, we can define universal learning machine equivalently below.

\begin{definition}[\bf  Universal Learning Machine (by Black Set)] 
For a $N$-1 learning $\mathcal{M}$, if its current black set is $B$, and a given objective pattern $p$, $\mathcal{M}$ can start from $B$ to learn and at the end of learning its black set becomes $p$, we call $\mathcal{M}$ can learn from $B$ to $p$. If for any $B$ and $p$,  $\mathcal{M}$ can learn from $B$ to $p$, we call $\mathcal{M}$ a universal $N$-1 learning machine.   
\end{definition}
For a $N$-1 learning $\mathcal{M}$, it is easy to see definition 4.1 and definition 2.2 are equivalent. 

Now, we turn attention to how to make a learning machine learn from $B$ to $p$. It is easy to imagine, there are many possible ways to learn. Here, we discuss learning by teaching, that is to say, we can design a special data sequence $T$ and apply it to the learning machine, then the machine learns  effectively under the driven of $T$. We call $T$ as teaching sequence. Teaching sequence is a specially designed data sequence.

It is easy to imagine, if we know the teaching sequence, learning by teaching is easy. Just feed the teaching sequence into, and learning is done. It is quite close to programming. But, learning by teaching can reveal interesting properties to us, and can guide us for further discussions. 

Consider a teaching sequence $T = \{(b_i, o_i) \ |\  i = 1, 2, \ldots\}$. Here, output feedback $o_i$ could be empty, i.e. there is just no output feedback. Learning machine still could learn without output feedback. Of course, with output feedback, the learning will be more effective and efficient. Teaching sequence is the only information source for the machine. Learning machine will not get any other outside information besides teaching sequence. This is very essential.

The fundamental question would be: what kind properties of learning machine to make it universal? We will reduce this questions to see some capabilities of learning machine, and with these capabilities, machine is universal.

Note, one special case is: black set of  $\mathcal{M}$ is empty set, we call it as empty state. This is one very useful case. There are some base patterns quite unique: $b_1 = (1, 0, \ldots, 0)$, $b_2 = (0, 1, \ldots, 0)$, $b_N = (0, 0, \ldots, 1)$, i.e. these base patterns only has one component equals 1, and rest equals 0. We call such base patterns as elementary base patterns.

\begin{definition}[\bf Learning by Teaching - Capability 1]
For a learning machine $\mathcal{M}$, the capability 1 is: for any elementary base pattern $b_j$, $j=1, 2, \ldots, N$, there is one teaching sequence $T_j, j= 1, 2, \ldots, N$, so that starting from empty state, driven by $T_j$, the black pattern become $b_j$.
\end{definition}
The capability 1 means: $\mathcal{M}$ can learn any elementary base pattern  from empty state.

\begin{definition}[\bf Learning by Teaching - Capability 2]
For a learning machine $\mathcal{M}$, the capability 2 is: for any black pattern $p$, there is at least one teaching sequences $T$, so that starting from $p$, driven by $T$, the black set becomes empty.
\end{definition}
The capability 2 means: to forget current black pattern, can go back to empty state.

\begin{definition}[\bf Learning by Teaching - Capability 3]
For a learning machine $\mathcal{M}$, the capability 3 is: for any 2 objective patterns $p_1$ and $p_2$, there is at least one teaching sequence $T_d$, so that starting from $p_1$, driven by $T_d$, the black pattern becomes $p_1 \cdot p_2$; and there is at least one teaching sequence
$T_p$ so that starting from $p_1$, driven by $T_p$, the black pattern becomes $p_1 + p_2$; and there is at least one teaching sequence
$T_n$ so that starting from $p_1$, driven by $T_n$, the black pattern becomes $\neg p_1$;
\end{definition}
Simply say, capability 3 means: for any 2 objective patterns $p_1, p_2$, learning machine is capable to learn subjective pattern of applying operator "$\cdot$", "+" to $p_1$ and $p_2$, and "$\neg$" to $p_1$. This is the most crucial capability. 

If one learning machine has all 3 capabilities, we expect a strong learning machine. Actually, we have following major theorem.

\begin{theorem}
If a $N$-1 learning machine $\mathcal{M}$ has the above 3 capabilities, it is an universal learning machine. 
\end{theorem}
{\bf Proof: } Since we have capability 2, we only need to consider the case: to start from empty state. That is to say, we only need to prove this: for any objective pattern $p$, we can find a teaching sequence $T$, so that starting from empty state, driven by $T$, the black pattern becomes $p$. 

According to Theorem 4, for any objective pattern $p$, we can find an X-form $E(b)$, where $E$ is one algebraic expression, $b$ is a group of elementary base patterns $b = \{e_i | i = 1, \ldots, K, K \le N\}$, so that $p$ equals this X-form, i.e. $p = E(e_1, e_2, \ldots, e_K)$. 

By $E$, we can construct teaching sequence like this way: \\
1) First we have a teaching sequence $T_1$ so that $\mathcal{M}$ go to empty state. This is using capability 1.\\
2) Then, have a teaching sequence $T_2$ so that $\mathcal{M}$ have black pattern $e_1$. This is using capability 2. \\
3) Since $E$ is formed by finite steps of $\cdot$, $\neg$, and $+$ starting from $e_1$, we can use capability 3 consecutively to construct teaching sequence $T$ for each operator in $E$. Eventually, we will get a teaching sequence over all operators in $E$.\\
Such teaching sequence $T_1 + T_2 + T$ will drive $\mathcal{M}$ to $p$.   

Note: The expression $E$ depends on several things: the complexity of $p$, and to find an X-form $E$ for $p$. In theorem 4, we demonstrated 2 level X-forms. We actually expect to have a much better X-form. The worst case would be: $E = b_1 + b_2 + \ldots$, in which, the pattern $p$ is so complicated that there is no way to find a X-form for higher level, so the only way is to just list all base patterns.

\begin{corollary}
If we have $N$-1 learning machine $\mathcal{M}$ with the above 3 capabilities, we then can use it to build one universal $N$-$M$ learning machine.
\end{corollary}

This is just following Theorem 5 and Theorem 2. From Theorem 5 and corollary, we reduce the task to find university learning machine to find a $N$-1 learning machine with 3 capabilities. Once we can find a way to construct a $N$-1 learning machine with those 3 capabilities, we have an universal learning machine. 

Also, it is easy to see that an universal learning machine surely has the 3 capabilities. Thus, the necessary and sufficient conditions for a learning machine to become universal are the 3 capabilities. 

But, do we have one learning machine with those 3 capabilities? Well, it is up for us to design a concrete learning machine with the 3 capacities. We will do this in other places. Any way, the 3 capabilities will give us a clear guide on design of effective learning machine: The most essential capabilities for a learning machine is to find a way to move patterns to higher organized patterns. See the quotation at the front, most important step is: "from a lower level to ...... higher". This indeed guides us well.

\section{Learning without Teaching Sequence}\label{table}
Learning by teaching is a very special way to drive learning. From discussions in last section, we can see clearly,  only when we have full knowledge of learning machine and the desired pattern, we could possibly design a teaching sequence. In this sense, learning by teaching is quite similar to programming -- to inject the ability into the machine, not machine to learn by itself. Of course, learning by teaching is still a further step than programming, and it will bring us a lot more power to handle machines than just programming. 

We focus on $N$-1 learning machine $\mathcal{M}$. 

{\bf Typical Mechanical Learning } \\
From examples of mechanical learning, typical mechanical learning would be as below: 
\begin{enumerate} [topsep=0pt,itemsep=-1ex,partopsep=1ex,parsep=1ex]
\item For $N$-1 learning machine $\mathcal{M}$, the learning target is often is given as an objective pattern $p_o$, $\mathcal{M}$ is expected to learn, and the learning result is that the black set of $\mathcal{M}$ become $p_o$. 
\item To drive the mechanical learning, data sequence is fed into $\mathcal{M}$. In learning by teaching, the data sequence is a specially designed teaching sequence. In learning without teaching, typically, data to feed into $\mathcal{M}$ are chosen from target objective pattern $p_o$, and from $p_o^c$. In another word, it is sampling $p_o$.
\item Feed-in data will drive learning, i.e. the black set of $\mathcal{M}$ is changing. Hopefully, at some moment later, the black set $B_t$ at the moment $t$ becomes $p_o$, or at least $B_t$ approximates $p_o$ well. 
\end{enumerate}

We put the above observations into a formal definition.

\begin{definition}[\bf Typical Mechanical Learning]
Let $\mathcal{M}$ be a $N$-1 learning machine, action of typical mechanical learning is: 
\begin{enumerate} [topsep=0pt,itemsep=-1ex,partopsep=1ex,parsep=1ex]
\item to set one target pattern: $p_o \subset PS_0^N$; 
\item to choose one sampling set $S_{in} \subset p_o$, normally, $S_{in}$ is a much smaller set than $p_o$. But, in extreme case, could be $S_{in} = p_o$; 
\item to choose another sampling set $S_{out} \subset p_o^c$, i.e. all member in $S_{out}$ is not in $p_o$. $S_{out}$ is a much smaller set than $p_o^c$. But, in extreme case, could be $S_{out} = p_o^c$; 
\item to use sampling set of $S_{in}$ and $S_{out}$ to form data sequence. In data sequence, data are $(b_i, o_i), i = 1, 2, \ldots$, if $b_i \in S_{in}$, $o_i$ is 1 or $\varnothing$ (empty), if $b_i \in S_{out}$, $o_i$ is 0 or $\varnothing$ (empty). 
\item to feed data sequence into $\mathcal{M}$ consecutively, we do not restrict how to feed, and how long to feed, and how often to feed, how to repeat feeding, which part to feed, etc.
\end{enumerate}
The action above will drive $\mathcal{M}$ to learn. As the result of learning. its processing (equivalently, black set) is changing.
\end{definition}
Remark: $S_{out}$ could be empty, i.e. not sampling out of $p_o$. But, $S_{in}$ is often not empty. However, if $S_{in}$ is empty, $S_{out}$ should not be empty. We will discuss this more in Data Sufficiency. 
 
For such typical mechanical learning, what is happening in the learning process? To address this, first we want to examine learning machine.
\bigskip

{\bf Internal Representation Space }\\
For a learning machine $\mathcal{M}$, it has input space ($N$-dim binary array), and output space ($M$-dim binary array, but here $M=1$), and something between input space and output space. This something between is the major body of a learning machine, and we denote it as $\mathcal{E}$. What is $\mathcal{E}$? We have not discussed it yet. We need to carefully describe $\mathcal{E}$ and its essential properties.

At any point of learning, if we stop learning, then $\mathcal{M}$ is a IPU, i.e. it has processing $F: PS^N_0 \to \{0, 1\}$ at the moment. So we can say, at this moment, $F$ uniquely defines something between input and output. Thus, at the moment, we can think, between input space and output space is $F$. Thus, it is quite reasonable to define $\mathcal{E}$ as the collection of all processing of $\mathcal{M}$. And, we will give a better name to $\mathcal{E}$: Internal Representation Space. 

\begin{definition}[\bf Internal Representation Space]
For $N$-1 learning machine $\mathcal{M}$, the major body of $\mathcal{M}$ that lays between input space and output space is called as internal representation space of  $\mathcal{M}$. At any moment, the processing of $\mathcal{M}$ is one member of this internal representation space. So, the internal representation space is the collection of all possible processing of $\mathcal{M}$. We denote it as $\mathcal{E}$. 
\end{definition}
Remark: All possible processing of $\mathcal{E}$ is $2^{2^N}$, an extremely huge number for not too small $N$. But for a particular learning machine, its internal representation space might be limited, not fully. 

For $N$-1 learning machine, for any processing $F$, it is equivalent to its black set $B$. By theorem 4, there is at least one X-form (one algebraic expression $E$, and some base patterns $b_1, b_2, \ldots, b_K$) so that $B = E(b_1, b_2, \ldots, b_K)$. We say that this X-form expresses processing $F$. Thus, naturally, we can think, the collection of all X-forms can be used to express the internal expression space. We have following definition.

\begin{definition}[\bf Internal Representation Space (X-form)]
For $N$-1 learning machine $\mathcal{M}$, the major body of $\mathcal{M}$ that lays between input space and output space is called as internal representation space of  $\mathcal{M}$. At any moment, one X-form expresses the processing of $\mathcal{M}$, it is one member of this internal representation space. So, the internal representation space is the collection of all possible X-forms. We denote it as $\mathcal{E}_X$.
\end{definition}
Remark, for one processing (which is equivalent to one black set), there is at least one X-form to express it.
Quite often, there are many X-forms to express one processing. So, the size of $\mathcal{E}_X$ would be not less than the size of $\mathcal{E}$. In fact, it is much larger. Learning sure is to get correct processing. However, to seek a good X-form that expresses the processing is more important. Thus, to use definition 5.3 (all X-forms as the internal representation space) is much better than to use definition 5.2. From now on, we will use definition 5.3. And, we just denote internal representation space as $\mathcal{E}$.

Now, we can clearly say, learning is a dynamics on space $\mathcal{E}$, from one X-form to another X-form. Or, we can say, learning is a flow on internal representation space. 

One important note: No matter what a learning machine really is, if it satisfies the definition of learning machine, it must have internal representation space as we defined above. If we concretely design a learning machine,  the internal representation space is designed by us explicitly, we know it well and can view its inside directly. If the learning machine is formed by different way, such as from a RBM (see Appendix), we could not view the inside directly. But, in theory, internal representation space indeed exists, and this space, equivalently, consists of a collection of X-forms. In theory, such space might be limited, not all X-forms, but only a part of the collection of all possible X-forms. This is not good. But, unfortunately, many learning machines are just so. However, when we discuss learning machine theoretically, the internal representation space is as definition 5.3.  
\bigskip

{\bf Learning Methods} \\
For a learning machine $\mathcal{M}$, besides input space, output space, and internal representation space $\mathcal{E}$, clearly, it must also have learning mechanism, or learning methods. So, we need to describe learning methods.

Now we know that learning is a dynamics on internal representation space, moving from one X-form to another. But, how exactly?

Let's make some notations. We have learning machine $\mathcal{M}$, its input space, output space, and its internal representation space $\mathcal{E}$, and a learning method $LM$. As in definition 5.1, we also have target pattern $p_o$, and data sequences $\{(b_i, o_i) \ | \ i = 1,2,\ldots\}$. Also assume the initial internal representation (one X-form)  is $e_0 \in \mathcal{E}$. 

Now, we start learning. First, one base pattern $b_1 \in S$ is feed into input space, and its feed-back value $o_1$ is also feed into output space ($o_1$ could be $\varnothing$ (empty), in that case, just has no feed-in to output space). Driven by this data, learning method $LM$ moves internal representation from $e_0$ to $e_1$, which can be written as:
$$
e_1 = LM(e_0, b_1, o_1 ) 
$$
Here, $LM$ is the learning method. Note, since the learning is mechanical, it is legible to write function form (if it is not mechanical, might not be justifiable to write in such function form). This is just the first step of learning. Next, we have: $e_2 = LM(e_1, b_2, o_2) = LM(e_0, b_1, o_1,  b_2, o_2)$. The process continues, we feed data  $(b_1, o_1), (b_2, o_2), \ldots, (b_k, o_k)$ into input space consecutively, and we have: 
\[
e_k = LM(e_0, b_1, o_1, b_2, o_2, \ldots, b_k, o_k ), k = 1, 2, \ldots  \label{eq:lmk} \tag{lm}  
\]
Note, the feed-in data could be repeating, i.e. could have $b_i = b_j$ while $i \ne j$.

This equation  \eqref{eq:lmk}  is actually the mathematical formulation for definition 5.1 -- typical mechanical learning.  

With this process, as $k$ increase, X-form $e_k$ continues to change, and we hope at some point, $e_k$ would be good enough for us. What is good enough? Perhaps, there are more than one criteria. For example, "$e_k$ to express $p_o$", i.e. the black set of $e_k$ equals the target pattern $p_o$. But, also, could be: "$e_k$ to express a good approximation to $p_o$". Or, additional to "express $p_o$, some additional goals are posted, such as $e_k$ is based upon less base patterns, etc.

Yet, how do we know $e_k$ would make our hope become true? Several questions immediately pop up:
\begin{enumerate} [topsep=0pt,itemsep=-1ex,partopsep=1ex,parsep=1ex]
\item What is the mechanism of $LM$ to make the $e_k$ approach $p_o$?
\item Is data sequence good enough? how to know data sequence is good enough? 
\end{enumerate}

We would first discuss sufficiency of data, then further discuss the learning mechanism. 
\bigskip

{\bf Data Sufficiency} \\
Learning machine needs data, and data drives learning. More data, more driving. But, data are expensive. It would be nice to use less data to do more, if possible. More importantly, we need to understand what data are used for what purpose.

As we know already, learning is actually to get one good X-form. But, one X-form normally is a quite complicated and is quite hard to get. How can a mechanical learning method get it? Mechanical learning is not as smart like human, it only follow certain simple and fixed rules. In order to make a mechanical learning to get a complicated X-form, sufficient data are necessary. But, what are sufficient data? Good thing is that X-form itself gives a good description of such data.

We already know that an X-form and all its sub-forms give perception bits. This tells us that X-form and all its sub-forms describe the structure of black set. To tell one X-form, the least data necessary are 2: one is in the black set, another is not in the black set. Of course, just 2 data is not sufficient to describe a X-form. However, how about for each sub-form, we can find such pair of data, one is in, and one is out? It turns out, all such pairs are very good description for the X-form. This is why we have following definitions.

\begin{definition}[\bf Data Sufficient to Support a X-form]
Suppose $e$ is a X-form, and suppose all sub-forms of $e$ are: $e_1, e_2, \ldots, e_L$. For a set of base patterns $DS \subset PS_N^0$, if for any sub-form $e_j$, there is at least one base patterns $b_j \in DS$ so that $e_j(b_j) = 0, e(b_j) = 1$, we said data set $DS$ is sufficient to support X-form $e$. That is to say, for each sub-form $e_j$, we have a data $b_j$ that is in black set of $e$, but not in black set of $e_j$.
\end{definition}
When we do sampling as in definition 5.1, if the sampling includes data sufficient to support X-form $e$, then the data sequence $D = \{(b_i, o_i) |  i =1, 2, \ldots\}$ will have such property: for each sub-form $e_j, j = 1, \ldots, L$, there is at least one data $(b_i, 1)$ in $D$ so that $e_j(b_i) = 0$. For such a kind of data sequence $D$, we say the data sequence is sufficient to support X-form $e$.

Data sufficient to support means: for each sub-form of a X-form, there is at least one data to tell learning machine, {\it This is only a sub-form. It is good, but not good enough}. With such information, learning method could conduct learning further mechanically. 

Data sufficient to support a X-form is to provide information from inside a X-form. But, we also need information from outside a X-form. To do so, we will define data sufficient to bound a X-form. In order to say more easily, we make some terms first. For 2 X-form $e$ and $f$, if $b \in PS_N^0$ and $e(b) = 1$ implies $f(b) = 1$, we say $f$ is over $e$ (this is equivalent that the black set of $f$ is greater than the black set of $e$). For 2 X-form $e$ and $f$, if there is $b \in PS_N^0$ so that $e(b) = 0$ and $f(b) = 1$, we say $f$ is out boundary of $e$ (this is equivalent to say the black set of $f$ is not subset of the black set of $e$).

\begin{definition}[\bf Data Sufficient to Bound a X-form]
Suppose $e$ is a X-form, and suppose all sub-forms of $e$ are: $e_1, e_2, \ldots, e_L$. For one sub-form $e_j$, if for any X-form $f$ that is both over $e_j$ and out boundary of $e$, there is at least one $b \in DS$ so that $e(b) = 0, f(b) = 1$, we call this data set $DS$ as sufficient to bound $e$.   
\end{definition}
When we do sampling as in definition 5.1, if the sampling includes data sufficient to bound X-form $e$, then the data sequence $D = \{(b_i, o_i) |  i =1, 2, \ldots\}$ will have such property: for each X-form $e'$ that is both over $e_j$ and out boundary of $e$, there is at least one data $(b_i, 0)$ in $D$ so that $e'(b_i) = 1$. For such a kind of data sequence $D$, we say it is sufficient to bound X-form $e$.

Data sufficiency to bound means: for any X-form that is over a sub-form, and out of boundary of $e$, there is at least one data $b$ to tell learning machine, {\it This X-form is not good, it is out of boundary}. With such information, learning method could conduct learning further mechanically.

{\bf Examples of Data Sufficient to Support a X-form: } \\
1. $e = b_1 + b_2$ is one X-form. Its all sub-forms are $b_1$ and $b_2$. So, $\{(b_1, 1), b_2\}$ are data sufficient to support $e$. \\
2. $e = b_1 \cdot b_2$ is one X-form. $e$ has no sub-form. Data set $\{b'\}$ or  $\{b''\}$ or $\{b'''\}$ are all data sufficient to support $e$. \\
3. $e = b_1 + (b_1 \cdot b_2)$ is one X-form. Its all sub-forms are $b_1$ and $b_1 \cdot b_2$. Data set $\{ b_1, b' \}$ is data sufficient to support $e$. And, so do  $\{ b_1, b'' \}$.  
\bigskip

{\bf Learning Strategies and Learning Methods } \\
Again, learning is a dynamics of X-forms, from one X-form to another. X-form is complicated. How come such a dynamics reaches the desired X-form? Such dynamics is determined by learning methods, and learning strategies. We discussed learning methods above, which is described well in equation  \eqref{eq:lmk}. Learning methods have set of rules on how to move from one X-form to another. Learning strategy is higher than learning method. It will govern these aspects: what X-forms to consider? what general approach to X-form? pre-set some X-forms? Or everything from scratch? etc. So, we can see that strategy governs method. Also, different strategy works for different kind of data. Different strategy also need different learning capabilities

We should emphasis here: learning is a complicated thing, one strategy and one method cannot fit all situations. There must be many strategies and even more methods. We are going to discuss some strategies and methods. But, still, there should have some common rules for these strategies and methods.

One very important property of X-form is: one processing (equivalently one black set) could be expressed by more than one X-form (normally, many). This property will play very important role in learning. Let's see one simple example first. Consider a set of base patterns $B$:
\[
B = \{ b_1, b_2, \ldots, b_K \}    \label{eq:bset} \tag{bs}
\]
$B$ has totally $K$ base patterns. What X-form could express $B$? The easiest one is:
\[
e = b_1 + b_2 + \ldots + b_k   \label{eq:exp} \tag{exp}
\]
Sure $e$ is one X-form to express $B$. Now, if we assume we can write $b_3, \ldots, b_K$ as some subjective expressions of $b_1$ and $b_2$, as following:
\[
b_3 = E_3 (b_1, b_2), \ldots, b_K = E_K (b_1, b_2)
\]
So, we can further have:
\[
e' = E_3 (b_1, b_2) + \ldots + E_K (b_1, b_2) = E' (b_1, b_2)  \label{eq:exp2} \tag{exp2}
\]
We can see X-form $e$ and $e'$ express the same black set. But, the 2 X-forms are very different. In fact, $e'$ is more complicated than $e$, and with higher structure. But at the same time, $e'$ is upon much less base patterns, just $b_1$ and $b_2$, while $e$ is upon on $K$ base patterns. 

This is very crucial: to learn $e$, we might have to use all base patterns $b_1, \ldots, b_K$, while to learn $e'$, in principle, we might only use 2 base patterns $b_1, b_2$ (just might, might need more, depends on learning method). And, not only that, it is much more. $e$ is just a collection of some base patterns, and no relationship between these base patterns are found and used, while $e'$ is built on many the relationship between base patterns (of course subjectively). In this sense, comparing to $e$, $e'$ is a better X-form to express black set $B$.  

The lesson to us is: to express one processing (or black set), there are many possible X-forms. Some such X-forms are simple, but, not good. Some of such X-forms are complicated, but, actually more robust. All learning strategies will use this fact.

One learning strategy will put some requirements on data and on learning machine. That is to say, data must strong to some point. And, learning machine is required to have certain capabilities. We do not specifically design a learning machine here. So, we do not know exactly how to realize such capabilities. But, we describe learning machine, and we show that with such capabilities, this strategy will work.

Now, we would propose some learning strategies. 
\bigskip

{\bf Strategy 1 -- Embed X-forms into Parameter Space}\\
This strategy will embed X-forms into parameter space, and use the dynamics on parameter space to drive the flow of X-forms. Parameter space $\mathbb{R}^U$ is a real Euclidean space, usually $U$ is a big number. In this strategy, we choose $L$ X-forms, $e_i, i = 1, 2, \ldots, L$ (so we only use some X-forms), and we will cut $\mathbb{R}^U$ into $L$ pieces, each piece is a region, denote as $V_i, i = 1, 2, \ldots, L$, so that $\mathbb{R}^U = \bigcup_{i=1}^L V_i$, then we associate each X-form with each region, $e_i \sim V_i$. In this way, we embed $L$ X-forms into the parameter space  $\mathbb{R}^U$. If we introduce a dynamics on $\mathbb{R}^U$, we actually introduce a dynamics on those $L$ X-forms. Since dynamics on $\mathbb{R}^U$ is a very familiar and mature mathematical topic, we have a great lot of tools to handle such dynamics. In this way, we can transfer the dynamics on X-forms into dynamics on $\mathbb{R}^U$. Or, we transfer learning to a dynamics on $\mathbb{R}^U$.

More exactly, we can write down this strategy as following. Let $\mathbb{R}^U$ be a real Euclidean space, $U$ is a big integer. And, by some way, $\mathbb{R}^U$ is cut into $L$ regions so that $\mathbb{R}^U = \bigcup_{i=1}^L V_i$, and  $V_i \cap V_{i'} = \varnothing$ for any $i, i'$. We also choose $L$ X-forms $e_i, i = 1, 2, \ldots, L$, and assign each $e_i$ to a region $V_i$. That is to say, for any $x \in \mathbb{R}^U$,  if $x \in V_i$, then on $x$, the X-form is $e_i$. We can denote this X-form associated with $x$ as $e_x$ as well, so $e_x = e_i$.  
We have data sequence $D = \{ (b_j, o_j) \ | j = 1, 2, \ldots, J\}$.  We also have one dynamics on $\mathbb{R}^U$ driven by this data sequence. We assume this dynamics is discrete. 

\[
x_k = LM(x_0, b_1, o_1, b_2, o_2, \ldots, b_k, o_k ), k = 1, 2, \ldots  \label{eq:embk} \tag{emb}  
\]
where $x_0$ is the initial point, and $x_k$ is the point at $k$-th step, $LM$ is the mechanism of dynamics. This equation is very similar to previous equation \eqref{eq:lmk}. For each $x_k$, we have the associated X-form $e_k$, $e_k \in \{e_1, e_2, \ldots e_L\}$. Thus, we see the learning is going on. 

Of course, we want learning to reach desired result. But, with what conditions and requirements, the above dynamics will reach the desired result? We have following theorem. We define a function $Lo$ on $\mathbb{R}^U$ as below:
\[
Lo(x) = \sum_{j=1}^J ( e_x(b_j) - o_j )^2,  \ \ \forall x \in \mathbb{R}^U  
\]
where $e_x$ is the X-form associated with point $x$.

\begin{theorem}
Suppose we have a desired X-form $e^*$ among those X-form we chosen $\{e_1, e_2, \ldots e_L\}$ , and we have a data sequence $D$ that is sufficient to support $e^*$ and sufficient to bound $e^*$, and the dynamics in equation \eqref{eq:embk} will reach the minimization target, i.e. as $k$ increases, $Lo(x_k)$ has trend to decrease, and at some $k$, $Lo(x_k)$ reach the minimum. Then, at the time $Lo(x_k)$ reaches minimum, we get the desired X-form, i.e. $e_x = e^*$.        
\end{theorem}
{\bf Proof:} The proof is quite clear. Since $D$ is sufficient to support $e^*$ and sufficient to bound $e^*$, for any point $x \in \mathbb{R}^U$, assume the associated X-form for $x$ is $e_k$, if $e_k$ is not $e^*$, then must have some $j$ so that  $e_k(b_j) \not= o_j$, so $Lo(x) > 0$. And, for $e^*$, for all $j$ $e^*(b_j) = o_j$. So $Lo(x)$ reaches minimum when the X-form associated with $x$ is actually $e^*$. 

So, by this strategy, we indeed find a way to learn: To design a dynamics on $\mathbb{R}^U$, which will reach the minimum of function $Lo$. For this strategy, we need strong data sequence that is both sufficient to support and to bound the desired X-form. 

Of course, there are some very critical issues to consider: 1) How to choose those X-forms $\{e_1, e_2, \ldots e_L\}$? 2) How to cut $\mathbb{R}^U$ into regions? 3) How to design a good dynamics $LM$? All these are big issues and not easy to deal. However, there should have many ways to choose, to cut, to design. 

One very important example for this strategy is deep learning. We can see in Appendix that deep learning is under this strategy. It is good to know that this strategy is a working and is currently produce many good results. 
\bigskip

{\bf Strategy 2 -- Squeeze X-form from Inside to Higher Abstraction}\\
Strategy 1 is to choose a good X-form from a previously given set of X-forms. Now, we will see another strategy, which builds the desired X-form from bottom up. We summarize this strategy as: 
\begin{enumerate} [topsep=0pt,itemsep=-1ex,partopsep=1ex,parsep=1ex]
\item Check input, to see if need to include the input. If so, add it. 
\item Squeeze current X-form to higher abstraction and more generalization, but not go over.
\item Choose best X-form from its internal representation space. 
\end{enumerate}
Learning will make sure X-form monotonously increase (in the term of its black set). This strategy require data to be sufficient to support the desired X-form. Note, no requirement on sufficient to bound. This is a huge difference. Also, this strategy needs machine has certain capability. We are not designing machine here. Here we just assume such capabilities, then to see what it can do.

\begin{definition}[\bf Strategy 2 - Capability 1]
Capable to squeeze current X-form $e$ to another X-form $e'$ with higher abstraction and more generalization. More precisely, the squeezing action will do following: Assume X-form $e = E(g)$, where $E$ is one algebraic expression (of 3 operators), and $g = \{b_1, \ldots, b_K\}$ is a set of base patterns, then X-form $e' = E'(g')$ should satisfy that $g' \subset g$, and $B \subset B'$, where $B$ is the black set of $e$, $B'$ is the black set of $e'$. If could find such a X-form $e'$, $e'$ is put it into internal representation space, otherwise, take no action.  
\end{definition}
There is no restrictions on how to do this squeeze (could be smart, or could be dumb). This capability can be simply said as: learning method will try to find a better organized X-form to replace current X-form, with the condition: its black set should be larger, not smaller. More generalization follows getting higher abstraction and bigger black set.

\begin{definition}[\bf Strategy 2 - Capability 2]
Capable to check a X-form $e$ to tell if $e$ is over the desired X-form or not, i.e. to tell if the black set of $e$ is a subset of the black set of the desired X-form.  
\end{definition}

Now, with the strategy summarized above, and with the capabilities and required data, we see what learning machine can learn.

\begin{theorem}
Suppose a learning machine $\mathcal{M}$ is using strategy 2 to learn, and it has Strategy 2 - Capability 1 and Capability 2, for a given black set $B_o$, if $e_o$ is a X-form that expresses $B_o$, and there is a data sequence $D = \{ (b_j, o_j) \ | j = 1, 2, \ldots \}$ that is inside $B_o$ and sufficient to support X-form $e_o$, then, start from empty X-form, if data in $D$ is fed to $\mathcal{M}$ fully and long enough (could be repeating feed, but not miss any data), eventually, $\mathcal{M}$ will learn $B_o$, i.e. the black set of $\mathcal{M}$ will become $B_o$.        
\end{theorem}
{\bf Proof:} We first describe the learning action as each data feed in. Suppose current X-form is $e$, and data feed in is $(b, o)$. Since data are all inside $B_o$, $o = 1$. Then learning method will check $e(b)$, if $e(b) = 1$, no need to do more, move to next data; if $e(b) = 0$, then this $b$ needs to be included, so replace $e$ with $e + b$. Then, capability 1 is used to get a X-form $e'$ with higher abstraction. Then, capability 2 is used to check if $e'$ is beyond the desired X-form. If $e'$ is OK, this X-form is put into internal representation space, and $e'$ replaces current X-form $e$. This is one step how learning is conducted.

Starting from $e = \varnothing$. The first data is $(b_1, 1)$. $b_1$ is a base pattern. This case, sure $e' = b_1$, and put in internal representation space. Since it is just beginning, this step is done. So, X-form becomes $e_1 = e' = b_1$. Note, at this time, the black set of current X-form is $B_1 = \{b_1\}$, so $B_1 \subset B_o$. 

This process continues. Now, consider step $k$. This time, current X-form is $e_{k-1}$. Input is $(b_k, 1)$. It will decide if $e' = e_{k-1}$ or $e' = e_{k-1} + b_k$. The logic here is: if $e_{k-1}(b_k) = 0$ , it means that $e_{k-1}$ is not good enough, it should be expand, so $e' = e_{k-1} + b_k$; if $e_{k-1}(b_k) = 1$, no need to expand, so, $e' = e_{k-1}$. 

Then, capability 1 is exercised. $e'$ is squeezed into higher abstraction, and the generalization is done at the same time. The result is a new X-form $e''$. Note, the capability 1 will make the black set of $e''$ getting bigger, at least not smaller. Moreover, capability 2 is exercised to check if $e''$ is out bound. If not, then set $e_k = e''$, i.e. update the X-form, and put $e''$ into internal representation space. 

This is the learning. We want to show, as $k$ increases, eventually, the black set of $e_k$ will become $B_o$. Since we know that the black set of $e_k$ is always a subset of $B_o$, so we only need to show that as $k$ increases, the black set of $e_k$ will not stay as a true subset of $B_o$ and not expand. But, this is clear. If at some $k$, the black set of $e_k$ is a true subset of $B_o$, so $e_k$ is sub-form of the X-form. Since data $D$ is sufficient to support the X-form, there must be a $k' > k$, at $k'$ data is $(b_{k'}, 1)$ and $e_{k'}(b_{k'}) = 0$. So, according to the learning process, $e_{k'}$ will be expanded. Proof is done.
\bigskip

We add another capability: Capability to forget current expression. That is to say, there are some special data, driving by them, learning machine could forget current X-form and make its X-form become empty. We name this capability as Capability Going Empty.

\begin{corollary}
A learning machine with Strategy 2 - Capability 1 and Capability 2 and Capability Going Empty is an universal learning machine.
\end{corollary}
{\bf Proof:} For any given black set $B_o$, we can find one X-form $e$ so that $B_o$ is black set of $e$. The data set $S \subset B_o$ so that it is sufficient to support $e$, according to above theorem, can be used to drive learning machine learn $e$ from any empty. Such data $S$ sufficient to support $e$ indeed exists. So, this machine is universal.
\bigskip

{\bf Strategy 3 -- Squeeze X-form from Inside and Outside to Higher Abstraction}\\
This strategy is quite similar to Strategy 2. They can be thought as one. Only for the purpose to show how data and learning method should work together, we write them differently here. The differences are 1) In Strategy 3, data are both sufficient to support and sufficient to bound, but in Strategy 2, only sufficient to support. 2) In Strategy 2, we have  Capability 2, but in Strategy 3, no such capability. That is to say, Strategy 3 uses much stronger data, but need much less capability. 

\begin{definition}[\bf Strategy 3 - Capability 1]
Same as Strategy 3 - Capability 1.
\end{definition}

\begin{theorem}
Suppose a learning machine $\mathcal{M}$ is using strategy 3 to learn, and it has Strategy 3 - Capability 1, for a given black set $B_o$, if $e_o$ is a X-form that expresses $B_o$, and there is a data sequence $D = \{ (b_j, o_j) \ | j = 1, 2, \ldots \}$ that is sufficient to support and sufficient to bound X-form $e_o$, then, start from empty expression, if data in $D$ is fed to $\mathcal{M}$ fully and long enough (could be repeating feed, but not miss any data), eventually, $\mathcal{M}$ will learn $B_o$, i.e. the black set of $\mathcal{M}$ will become $B_o$.        
\end{theorem}
{\bf Proof:} Since this is quite similar as strategy 2, we only say the part different. The data are different. So, in this strategy, $b_k$ could be inside $B_o$ or outside $B_o$. When data $(b_k, o_k)$ feed in, possibly  1) $e_k(b_k) = 1, o_k = 1$, 2)  $e(b_k) = 0, o_k = 1$, 3) $e_k(b_k) = 0, o_k = 0$, 4)  $e(b_k) = 1, o_k = 0$. For case 1) and 3), no need to do anything, move to next data. For case 2), it means $b_k$ needs to be included, so expand to $e_k + b_k$. For case 4), it means $b_k$ should be excluded, so prohibit $b_k$, the way to do so: $\neg (e_k \cdot b_k) + (e_k  \cdot \neg b_k)$. Other than this part, the learning is same as in last theorem. 

Then, capability 1 is exercised. And, no capability 2. The proof is also similar.

\begin{corollary}
A learning machine with Strategy 3 - Capability 1 and Capability 2 and Capability Going Empty is an universal learning machine.
\end{corollary}

We briefly comment these 3 strategies below.
\begin{enumerate} [topsep=0pt,itemsep=-1ex,partopsep=1ex,parsep=1ex]
\item They put different requirements on data and on machine. Strategy 2 only require data sufficient to support. This is much less than both sufficient to support and to bound. But, strategy 2 requires machine to have a very strong capability. Strategy 3 and 1 put same requirements on data. However, strategy 1 put a super strong requirement on setup. The desired X-form must be set into the regions.  
\item We just say, the above 3 strategies are not all learning strategies. There are many other learning strategies and methods.
\item Human perhaps never learn anything from very scratch. Previous learning results are used to help new learning. All learning strategies should utilize this approach. This is a huge topic. We will discuss this issue later.
\end{enumerate}

\section{More Discussions about Learning Machine}\label{table}
{\bf Learning vs. Approximating}\\
Very often people are saying: "Machine learning is nothing but a kind of approximation to probability distribution". We would argue this view is quite far from reality. For the sake of arguments, let's see dictionary definition first. We use online dictionary "www.dictionary.com". For learning, the definition is "to acquire knowledge of or skill in by study, instruction, or experience"; for approximating, "to come near to; approach closely to; to estimate". They are very different. Of course, this pair of 2 words, like many other pairs, indeed have something in common, such as "to approach the knowledge vs. to acquire knowledge". It is hard to make one absolute black and white differentiation. However, we would like to point out one big difference: learning is to use outside information to build up one best possible inside model, while approximating is to use outside information to get closer to a model previously chosen. We would think this cutting line is acknowledged by most people. 

By this cutting line, we can see clearly that mechanical learning is indeed learning, not approximating, since it is using information of feed in data to get an X-form and drive X-form to become better. Mechanical learning is not approximating a model previously defined. In fact, there is no previously defined model at all. As we see in appendix, deep learning is not approximating either. Deep learning is trying to get a better X-form from a set of chosen X-forms (huge number of such X-forms) by using certain methods (like CD+SGD, etc). It is very different from approximating probability distribution.     

The above argument seems a little trivial. However, to clarify what is learning and what is approximating could help us to clean mind greatly. We want machine to build its own model for outside (as smartly as possible), and do not want machine just follow one model previously defined, which could not be suitable for all situations, no matter how good it is.
\bigskip

{\bf Internal Representation Space}\\
A learning machine consists of several parts. However, internal representation space is the most essential part. It might not be explicitly viewable, but it indeed exists. In our definition of learning machine, we do not specifically require a internal representation space. However, it comes as a consequence of being a learning machine. Such fact illustrates the importance of internal representation space even more clearly. In practice, a learning machine could have no a intentionally designed internal representation space. For example, RBM is such a learning machine. However, as we can see in Appendix, internal representation space indeed exists and play essential role in learning. Again, if a learning machine satisfies Definition 1, it must have an internal representation space, even though we might not know exactly how internal representation space works.

Though we might not know exactly how internal representation space works, we know that it is equivalent to a space consists of X-forms. X-form is a subjective way of a learning machine to organize its view about patterns. Learning machine learns from its input data and tries to adapt a better X-form. A better X-form means to have a better way to perceive and process the incoming pattern.

Thus, to have a good internal representation space would be the key to a successful learning machine. First, it should have sufficient expression power to include any necessary X-form; second, it should have good structure so that expressions can be easily understood; third, dynamics on X-form can be easily executed and easily understood. 

Here, we would like to raise such a question: In what way to mathematically describe internal representation space? We know it is collection of X-forms. This is actually one good mathematical description. However, can we do better? We need to find tools to describe relationship of X-forms and sub-forms in more details and to calculate X-form and sub-forms easily. One possible tool is catagory theory. This direction is very worth to pursue. 

We also point out one fact about internal representation space. As we see in the learning of both by teaching and without teaching, internal representation space serves a reservoir for X-forms. These X-forms, might not be useful for certain learning, however, could turn out to be very useful for others. Such fact highlights the importance of internal representation space. How to utilize this reservoir of X-forms is going to be critical for many applications. 
\bigskip

{\bf Combine 5 Learning Approaches together}\\
Petro Domingos in \cite{pedro} discussed 5 different learning approaches. In \cite{paper1}, we speculated possibilities to unite 5 different learning approaches. Now, we can confidently say: a good internal representation space indeed is the center to unite 5 different approaches together. We briefly put some thoughts below.

1. Logical Deduction Approach. 
This is very clear. One X-form is actually a logic statement. When doing learning, a lot of X-form would be accumulated in internal representation space. So, logical searching and logical deduction would be very naturally done on these X-forms. In this way, it is very natural to connect the learning machine with a set of logical deduction rules to generate new X-form, and compare new X-form with output feedback.  

2. Connectionist Approach.
Just looking into its definition, we can easily see that one X-form is a set of connections. So, it is already a connectionist approach. Reversely, for any connectionist model, all connections would form a visible internal representation space. Thus, to focus on internal representation space actually means connectionist approach. This is also a strong evidence that good learning machine should utilize connectionist model.

3. Probability Approach.
In principle, we can build probabilistic view on X-forms. For example, if we assign some probabilistic measure on X-forms, probabilistic operation can be done on learning. 
In this way, a Bayes like model can be build on the movement of X-forms. Also see "Deterministic vs. Probabilistic" below.

4. Analogy Approach. 
In internal representation space, we have many X-forms. Some X-form might be quite close to another, by this way or that way, subjectively or objectively, logically or probabilistically, realistically or imaginably, etc. So, analogy appears naturally. We can think analogy is just a different way to view those X-forms, transformed way, different connection way, distorted way, etc. Not only we can easily view analogy, we can also easily execute analogy, since X-form is already connect well to processing. For example, if we first have X-form $e$, and we can see another X-form $e'$, and see $e'$ as an analogy to $e$, we can immediately have execution of $e'$: just use $e'$ to replace $e$. It is very natural to conduct analogy approach with internal representation space. Also, analogy has a lot to do with subjectiveness. So, learning method utilizing analogy could be very efficient and powerful.

5. Evolution Approach.
If we stay with level 1 learning machine, there is no evolution, since learning method is not changing. But, if we go to level 2 learning machine, evolution starts immediately. We can go even higher level, i.e. evolution of evolution. See, our learning machine capture evolution well, and make all related execution easily realizable. 

Summarize, we now know that in a learning machine, if we have a good internal representation space, all kinds of learning approaches can live together and support each other on the internal representation space. This is one huge advantage. We can use each method for its own strength to get a better learning machine.
\bigskip

{\bf Deterministic vs. Probabilistic}\\
Our definition makes learning machine deterministic, and everything followed is also deterministic. In this framework, any probabilistic view, if any, is just supplementary. However, we would like to point out, 100\% deterministic is difficult. Probabilistic view indeed has advantages in some aspects, and it is necessary. For example, often it is easier to get a probability measure than an exact function. But, we would like to point out: learning machine should be mainly based on deterministic framework. There are compelling reasons for this. Input data indeed has rich intrinsic structure, and a learning machine needs to capture such structure in order to be efficiently and effectively learning and process patterns. If learning machine has 100\% probabilistic view, it would treat all things as a random event, its internal representation space would be very flat (i.e. no hierarchy structure), cause it very hard to get intrinsic structure. The reason to have a deterministic framework system is because the world around learning machine is structurally deterministic (data indeed has intrinsic structure). 

To adding probabilistic view, much more works are needs. Consider a $N$-1 IPU $\mathcal{M}$, and its processing $F: PS^0_N \rightarrow \{0, 1\}$. If we view it deterministically, $F$ is one mapping, which is exact and often hard to obtain. If we view it probabilistically, $F$ is not exact mapping, we can only consider probability of base pattern going to 1 or 0. Usually, it is much easier to approximate the probabilistic measure than to obtain exact mapping. However, easiness is only on surface. If we want to explore the inter-relationship of base patterns, such as "base pattern $b_1$ is base pattern $b_2$ and $b_3$ appears together with certain probability", it would be hard mathematically. This is why we only do deterministic view here. In this way, we hope to grasp the most essential understanding. 

However, our world is indeed full of probabilistic events, we do need to build one model combining intrinsic pattern structure with probability. In order to do so, we have to handle how logic ("+", "$\cdot$", "$\neg$", etc) propagate in probability measure, and how objective view and subjective view collides in probability space. Perhaps, this is a full view of new Bayes like logic. This is a great direction, we will try it in later works. Someone has done work from probabilistic view on deep learning, see \cite{gal}. However, the work is totally based on deep learning, which has no concept of subjectiveness and X-form. \\

{\bf Data Sufficiency} \\
We need data to drive learning. Generally speaking, more data, more learning. But, how much data are sufficient? Up to now, no any theory about this crucial issue exists. We use X-form and sub-form to understand this issue. We define data sufficient to support an X-form, and data sufficient to bound an X-form. Such sufficiency lays down a theoretical framework for us to understand data: why we need such amount of data for this learning? how much data are necessary for that learning? etc. But, to establish such data sufficiency is just beginning. 

In section 5, Strategy 2 and Strategy 3 show us that data sufficiency indeed depends on learning capability. Generally speaking, strong capability needs weaker data, and vise versa.  The relationship of data and learning strategy, methods and machine capability are very important. The data sufficiency we introduced is just the beginning, we need to more work in this direction. Also this is related to learning theory. 
\bigskip

{\bf Learning Strategies and Methods } \\
Learning is a dynamics on X-forms, from one X-form to another. Such a dynamics is determined by learning strategies and methods. A learning machine could be smart (i.e. it can learn better, using less data, learn faster, etc.), and could be dumb (more data, learn slow, etc.). Smartness or dumbness are mostly determined by learning strategies and methods. 

In section 5, we demonstrate 3 learning strategies. There should be many other learning strategies and methods. We invent strategy {\it squeezing to higher abstraction and more generalization} here. We should continue to invent more. 

Actually, learning method could be learned. We have defined Level 1 learning machine, which could not modify its learning method. Level 1 learning machine is what we focus on for most time. But, in many cases, it is good to introduce level 2 machine, which can modify its learning methods. In this sense, one approach should be: we are going to use learning to get more and better learning strategies and methods.
\bigskip

{\bf Mathematical Learning Theory} \\
If the learning machine is super intelligent, it might only need a few data to learn everything. But, we are discussing mechanical learning here. Learning machine only follow some simple and mechanical rules. It could not learn from nothing. Without sufficient data driven, it will not be able to learn. In order to understand these issues, it is best have a general theory about learning: by using what learning strategy and method, with what capabilities, by what kind of data, how much amount of data, what a learning machine can learn, with what kind of complexities. We strongly believe that such theory could be established, and such theory is mathematical. Though we cannot pursue such theory right now, we can speculate that this theory make learning clearer. This theory will be able to explain and measure the complexity of learning objects, structure of data, efficiency and effectiveness of learning methods and strategies, learning process and governing equations, and more. We might call this as Mathematical Learning Theory. Such a mathematical theory might be even expanded with philosophical flavor -- Mathematical Epistemology. This is one exciting view and we hope to do some concrete work in this direction later.
\bigskip

\section{How to Design Effective Learning Machine}\label{table}
There are 2 approaches to study learning machine: one is to well describe learning machine, another is to design one particular learning machine. Both approaches should support each other. In this work, we are doing the first approach. We will do the second approach in other place. But, since we get some good insights from work here, which serves as good guidance for us to design effective and efficient learning machine, we briefly discuss some issues of designing learning machine.

{\bf 1. It should have very effective and efficient internal representation space} \\
Without a good and easy to review internal represnetation space, a learning machine is hard to work well. How to make good internal representation space is not a simple issue. However, one thing is clear. It is not good to embed the internal representation space in some real parameter space $\mathbb{ R}^U$, where $U$ is a huge integer, like many millions. In that way, we will lose the ability to navigate and understand immediately. Armed with knowledge we discussed, we know that collection of X-forms is the internal representation space. It is naturally for us to find some mechanism to realize X-form in design. We also should realize that X-form is a connectionist model by its nature.

{\bf 2. It should be self-aware about new base patterns} \\
The self-awareness about new base patterns is one important property. If possible, we should make this property ready for a learning system. This needs big efforts to realize. 

{\bf 3. It is level 1 learning machine, but with ability to become higher level} \\
Learning method to be level 1 has advantage. It is simpler and eaiser to examine the learning. Since learning methods are not changing with the learning, if there is some trouble, it will be much easier to roll back. If it is not level 1, roll back would be much harder, or just impossible. However, it should be ready to move to level 2, or even higher learning level, if necessary. It should fully utilize evolution and inheritance methods if needed.

{\bf 4. It should have best possible prior knowledges} \\
In principle we want universal learning machine. But, in practice, one learning machine is for some special tasks. Often, we have prior knowledges about those tasks. To build these prior knowledges into learning machine could significantly improve learning. Prior knowledges could be build into learning methods and initial state of internal representation space.

{\bf 5. Allow more learning strategies and methods be available} \\
Learning methods should have abilities to conduct multiple reasoning (like deterministic, probabilitstic, etc). It should have more than one strategy and method. It can switch the strategies and methods by reasoning on feed-in data. Since we demonstrated in section 4 and 5 that with certain capabilities, a learning machine become universal, it is reasonable to make learning machine to have the capabilities of learning by teaching and learning without teaching. Very likely, they might be the minimal capabilities of an effective learning machine.

Guided by above principles, we have designed one type of learning machine: OSIPL machine. We will discuss details of OSIPL machine in other place.

\section*{Appendix}\label{table}
In the discussions of learning machine, we introduced our view on how mechanical learning is conducted. In this Appendix, we would like to use this view to see deep learning. Hopefully, from this special angle, we can see some interesting structure and details of deep learning. 

According to our definition, deep learning is mechanical learning, if without human intervene. Of course, this "if" is a big if. Often, deep learning program is running with a lot of human intervene. Actually, we introduce the term "mechanical learning" is for this purpose: to isolate a learning machine from human intervene so that we can discuss details of a learning machine. 

We would like to restrict our discussion to Hinton's original model \cite{hinton}, i.e., a stack of RBMs. Each level of RBM is clearly a $N$-$M$ learning machine ($N$ and$M$ are dimensions of input and output). Hinton's deep learning model is by stacking RBM together. If without further human intervene, it is a learning machine. This is the original model of deep learning. Other deep learning program can be thought as variations based on this model. Though in the past few years deep learning has leaped forward greatly, stacking RBM together is still the most typical deep learning model.  

Thus, we would expect many things we discussed could be applied to deep learning. The point is, we are viewing it from a quite different angle -- angle of mechanical learning. For example, we can view Hinton's original deep learning program \cite{hinton} as one 258-4 learning machine, and we ask what is the internal representation space of it? We expect such angle and questions would reveal useful things for us.

The central questions indeed are: what is the internal representation space of deep learning? what is the learning dynamics? At first, seems it is quite hard since learning is conducted on a huge parameter space, and learning methods are overwhelmingly a big body of math. However, when we apply the basic thoughts of learning machine to deep learning, starting from simplest RBM, i.e. 2-1 RBM, we start to see much more clearly.
\bigskip

{\bf 2-1 RBM} \\
2-1 RBM is the simplest. However, it is also very useful since we can examine all details, and such details will give us a good guide on more general RBM. 

2-1 RBM is one IPU. We know 2-1 IPU totally has 16 processing ($2^{2^2}$). But, we only consider those processing: $p(0, 0) = 0$, so totally 8 processing, which we denote as $P_j, j=0, \ldots, 7$ (see \cite{paper1}). For 2-1 RBM, any processing $p$ can be written as: for input $(i_1, i_2)$, the output $o$ is:
$$
o = p(i_1, i_2) = \text{Sign}(a i_1 + b  i_2), \text{where} \ (a, b) \in \mathbb{R}^2, \ \text{Sign}(x) =
\left\{
	\begin{array}{ll}
		1  & \mbox{if } x \ge 0 \\
		0  & \mbox{if } x < 0
	\end{array}
\right.
$$


The parameters $(a, b)$ determine what the processing really is. Parameter space $\mathbb{R}^2$ has infinite many choices of parameters. But, there are only 6 processing, thus, for many different parameters, the processing is actually same. We can see all processing in below table:

\begin{center}
\centering
    \begin{tabular}{|c|c|c|c|c|c|c|c|c|c|c|c|c|c|c|c|c|}
        \hline
        ~     & $P_0$ & $P_1$ & $P_2$ & $P_3$ & $P_4$ & $P_5$ & $P_6$ & $P_7$  \\ \hline
        (0,0) & 0     & 0     & 0     & 0     & 0     & 0     & 0     & 0       \\ 
        (1,0) & 0     & 1     & 0     & 1     & 0     & 1     & 0     & 1       \\ 
        (0,1) & 0     & 0     & 1     & 0     & 1     & 1     & 0     & 1       \\ 
        (1,1) & 0     & 0     & 0     & 1     & 1     & 0     & 1     & 1       \\
        \hline
          Region   & $R_4$ & $R_3$ & $R_5$ & $R_2$ & $R_6$ & None & None & $R_1$   \\	
        \hline
          X-form    & $0$ & $b_1$ & $b_2$ & $b_1 + b_3$ & $b_2 + b_3$ & $b_1 + b_2$ & $b_3$ & $b_1 + b_2 + b_3$   \\	
        \hline
    \end{tabular}

{\bf Tab. 1.  Table for all processing of 2-1 RBM} 
\end{center}


\begin{center}
\begin{picture}(220,250)(0,0)
\put(0, 10){\resizebox{9 cm}{!}{\includegraphics{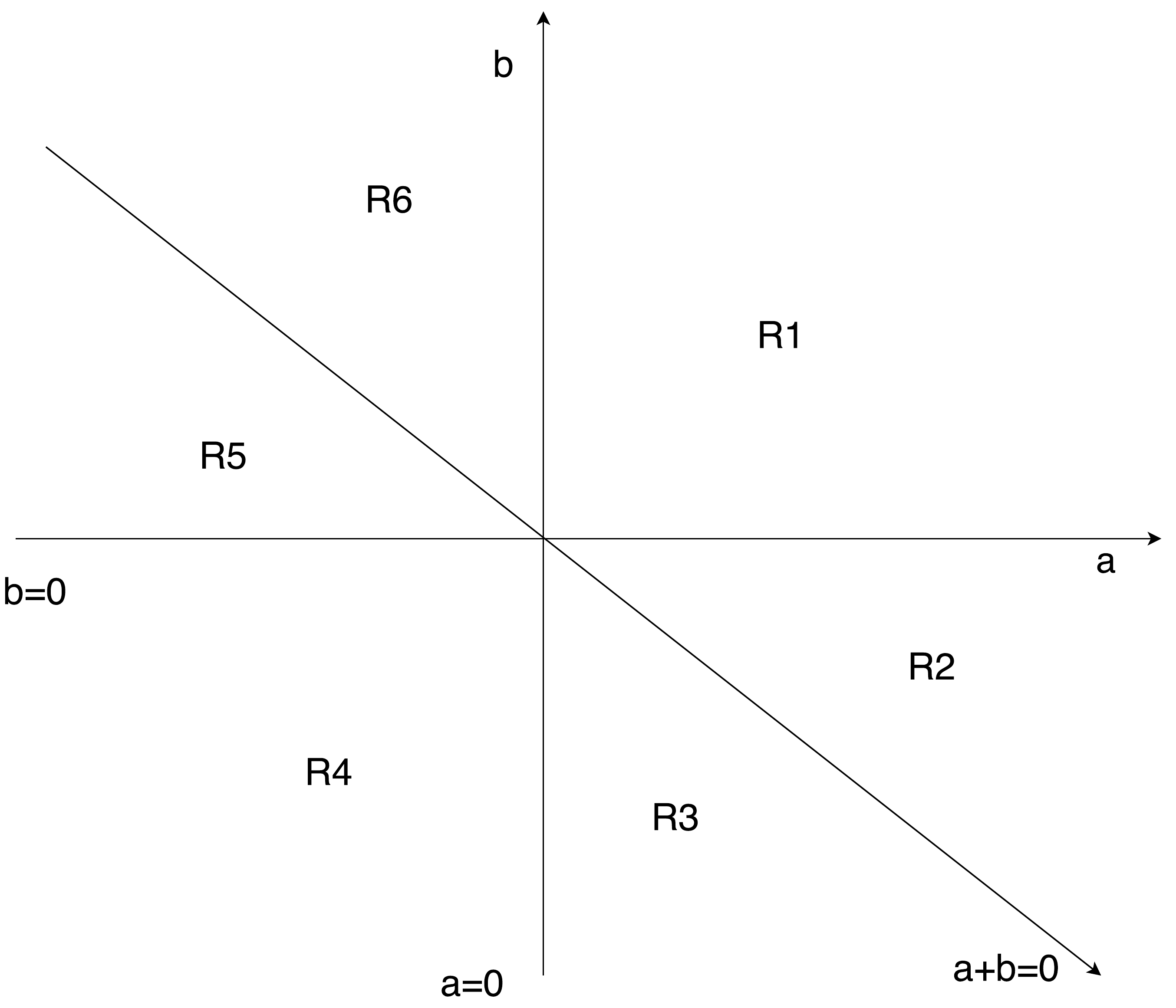}}}
\end{picture}

{\bf Fig. 1. Parameter space and regions of 2-1 RBM} 
\end{center}

In first row of table, $P_i, i = 0, \ldots, 7$ are all processing of 2-1 IPU. Under first row, there is value table for each processing. We point out some quite familiar processing: $P_7$ is actually logical OR gate, $P_6$ is logical AND gate, $P_5$ is logical XOR gate. Note, $P_5, P_6$ are processing for 2-1 IPU, but not in 2-1 RBM. It is well known, 2-1 RBM has no XOR and AND (i.e. no $P_5, P_6$).

$R_j$, $j=1, \ldots, 6$ indicate regions in parameter space $\mathbb{R}^2$, each region for one processing. There are only 6 regions, since 2-1 RBM only has 6 processing. We briefly discuss how we get these regions. See illustration in figure.

Suppose $p$ is processing. Line $a = 0$ cuts $\mathbb{R}^2$ into 2 regions: $a \ge 0$ and $a < 0$. If $(a, b)$ is in first region, then $p(1, 0) = 1$, in second, $p(1, 0) = 0$. Line $b = 0$ is perpendicular to $a=0$, so, it cuts the previous 2 regions into 4 regions: $a \ge 0, b \ge 0$ and $a \ge 0, b < 0$ and  $a < 0, b \ge 0$ and $a < 0, b < 0$. Clearly, if $b \ge 0$, $p(0, 1) = 1$, if $b < 0$, $p(0, 1) = 0$. Line $a + b = 0$ could no longer cuts the previous 4 regions into 8 regions, it could only cut 2 regions (2nd, 4th quadrant) into 4 regions ($R_2, R_3, R_5, R_6$). So, totally, we have 6 regions, and each region is for one processing. This argument about regions is very simple, yet very effective. We can extend this argument to $N$-1 RBM.
 
Each region is for one processing. So, region can be used to represent processing. That is 6th row in the table shown. Yet, a much better expression is by X-form (\cite{paper2}). We explain them here. 
Here $b_0 = (0, 0), b_1 = (1, 0), b_2 = (0, 1), b_3 = (1, 1)$ are base patterns. For 2-dim pattern space, there are only these 4 base patterns. But, $b_i$ can also be used to represent one processing of 2-1 IPU, i.e. $b_i$ is such a processing: when input is $b_i$, output is 1, otherwise output is zero. X-forms are expressions based on all $N$-1 base patterns, operations +, $\cdot$, $\neg$, composition, Cartesian product, and apply them consecutively. Example, $b_1 + b_3$, $b_1 \cdot b_2$, $b_1 + \neg b_2$ are X-forms. 
Any processing of 2-1 IPU can be expressed by at least one X-form \cite{paper2}. For example, if region is $R_3$, processing is $P_1$, X-form is $b_1$. 
Another example, region is $R_1$, processing is $P_7$ (this is OR gate), X-form is $b_1 + b_2 + b_3$. $P_5$ is a processing of 2-1 IPU (XOR gate), but not in 2-1 RBM, its X-form is $b_1 + b_2$.  
The 7th row in the table shows X-forms representing processing. We can say that each processing is represented by a region, and by a X-form as well.

When 2-1 RBM is learning, clearly, parameter $(a, b)$ is adapting. But, only when $(a, b)$ cross region, processing changes. Before crossing, change of parameters is just for preparation for crossing (perhaps many parameter changes are just wasted). Learning is moving from one region to another. Or, equivalently, learning is from one X-form to another. Such view is crucial. Now, we are clear, on surface, learning on 2-1 RBM is a dynamics on parameter space $\mathbb{R}^2$, but real learning dynamics is on 6 regions (or X-forms). Such indirectness causes a lot of problem. 
\bigskip

{\bf 3-1 RBM} \\
Just increase input dimension 1, we have 3-1 RBM. To discuss it, we can gain some insights for general RBM. For 3-1 RBM, still we can write: for any input $(i_1, i_3, i_3) \in \mathbb{B}^3$, output $o \in \mathbb{B}$ is:
$$
o = p(i_1, i_2, i_3) = \text{Sign}(a i_1 + b  i_2 + c i_3), \text{where} \ (a, b, c) \in \mathbb{R}^3, \ \text{Sign}(x) =
\left\{
	\begin{array}{ll}
		1  & \mbox{if } x \ge 0 \\
		0  & \mbox{if } x < 0
	\end{array}
\right.
$$

However, while we can easily write down all possible processing of 2-1 RBM, it would be hard to do so for 3-1 RBM. For 3-1 IPU, we know that the number of all possible processing is $2^{2^3} = 2^8 = 256$. Since only considering such processing $p$: $p(0, 0, 0) = 0$, the number becomes $256/2 = 128$. We expect 3-1 RBM has less processing. But, how many possible processing of 3-1 RBM could have?

Following the guidance that 2-1 RBM gives to us, i.e. to consider the hyperplane generated by nonlinearity that cuts parameter spaces, we examine parameter space $(a, b, c) \in \mathbb{R}^3$, and following planes:
$a = 0, b = 0, c = 0, a+b = 0, a+c = 0, b+c = 0, a+b+c =0$. 
These planes are naturally obtained. For example, if we consider the input $(1, 0, 0)$, it is easy to see plane $a = 0$ is where cut the value of output: 1 or 0. So, the above planes are associated with following inputs:
$(1, 0, 0), (0, 1, 0), (0, 0, 1), (1, 1, 0), (1, 0, 1), (0, 1, 1), (1, 1, 1)$

We can clearly see that in one region that is cut out by above 7 planes, the output values are same. Therefore, one region actually is representing one processing: in the region, processing is same. So, question of how many possible processing becomes how many possible regions cut out by the 7 planes. We do counting for the regions below. 

First, $a = 0$ cuts parameter space into 2 pieces: $R_1^1$, $R_2^1$. 
Second, $b = 0$ perpendicular to $a = 0$, so, it cuts each region  $R_1^1$, $R_2^1$ into 2 pieces, we then have 4 regions: $R_1^2$, $R_2^2$.$R_3^2$, $R_4^2$. 
Then, $c = 0$ perpendicular to $a = 0$ and $b = 0$, so, we have 8 regions: $R_j^3, j=1, \ldots, 8$.
Then, consider $a+b = 0$. This plane no longer could be perpendicular to all $a = 0, b = 0, c = 0$. We will not have $2 * 8 = 16$ regions. We only have $1.5 * 8 = 12$ regions.
Following the same argument, we have: For $a+c = 0$, $1.5 * 12 = 18$ regions. For $b+c = 0$, $1.5 * 18 = 27$ regions. For $a+b+c = 0$, $1.5 * 27 < 41$ regions. 

So, for 3-1 RBM, there are at most 41 possible processing, comparing to 128 possible processing of full 3-1 IPU. However, there are possibility that the number of processing is even less than 41, since among those regions, it is possible that 2 different regions give same processing. We do not consider these details here. 

Since regions can be represented by X-form, each processing 3-1 RBM can be represented by at least one X-form.  $b_1 = (1,0,0), b_2 = (0,1,1), b_3=(0,0,1), \ldots, b_7 = (1,1,1)$ are X-form for all base patterns. Examples, X-form $b_1 + b_2$ is in 3-1 RBM.  But, $b_1 \cdot b_2$ is not in 3-1 RBM. There are a lot of such X-form that is not in 3-1 RBM.

Learning dynamics on 3-1 RBM is also in such way: on surface, it is dynamics on $\mathbb{R}^3$, but real learning dynamics is on 41 regions (or X-forms). 
\bigskip

{\bf $N$-1 RBM} \\
The argument for 3-1 RBM can be extended to $N$-1 RBM (See details in \cite{paper2}). We consider hyperplanes and regions cut off by these hyperplanes. The number of these regions is less than:  $2^N 1.5^{2^N-N-1}$. Compare to the number of all processing of $N$-1 IPU, which is $2^{2^N - 1}$, easy to see, $N$-1 RBM has much less processing. This means that $N$-1 RBM could not express many processing. 

For $N$-1 RBM, still we can write: for any input $(i_1, \ldots, i_N) \in \mathbb{B}^N$, output $o \in \mathbb{B}$ is:
$$
o = p(i_1, \ldots, i_3) = \text{Sign}(a_1 i_1 + a_2  i_2 + \ldots + a_N i_3), \text{where} \ (a_1, \ldots, a_N) \in \mathbb{R}^N, \ \text{Sign}(x) =
\left\{
	\begin{array}{ll}
		1  & \mbox{if } x \ge 0 \\
		0  & \mbox{if } x < 0
	\end{array}
\right.
$$

\begin{equation}
a_1 = 0, \ldots, a_1+a_2 = 0, \ldots, \ a_1+a_2+a_3 =0 \ldots, \ \ldots, \ a_1+a_2+a_3+\ldots = 0, \ldots
\end{equation}
There are $ {N\choose 1} $ hyperplanes such as $a_1 = 0$;  $ {N\choose 2} $ hyperplanes such as $a_1+a_2 = 0$;
\ldots. We also have this: First $N$ hyperplanes will cut parameter space into $2^N$ regions. Then, later each hyperplanes will cut more regions by the rate of multiplying factor $1.5$. Thus, we can see the number of regions are:

$$
2^N * 1.5^{K_2} * 1.5 ^{K_3} * \ldots * 1.5^{K_N}
$$
where ${K_2} = {N\choose 2}$ is the number of hyperplanes such as $a+b = 0$, etc. 

And, we have the equation:

\begin{equation}
2^N = {N\choose 0} +  {N\choose 1} +  {N\choose 2} + \ldots +  {N\choose N}   
\end{equation}

So, 

\begin{equation}
K_2 + K_3 + \ldots + K_N = {N\choose 2} +  {N\choose 3} + \ldots +  {N\choose N} = 2^N - {N\choose 1} -{N\choose 0}    = 2^N - N -1
\end{equation}

Thus, the number of regions are

$$
2^N * 1.5^{2^N-N-1} = 2^N * ({\frac{3}{2}})^{2^N-N-1} 
$$

This is a very big number. Yet, compare to the total possible processing of full IPU, it is quite small. See their quotient:

\[
\frac{2^{2^N}}{2^N * ({\frac{3}{2}})^{2^N-N-1} }   = 2* (4/3)^{2^N - N - 1}
\]
It tells that full IPU has $f = 2* (4/3)^{2^N - N - 1}$ folds more processing comparing to RBM. This is huge difference. Say, just for $N = 10$, $f$ is more than 120 digits, i.e. the number of processing of full IPU would has more 120 digits at the end.

Also, each region can be expressed by at least one X-form. For examples, $b_1 + b_N$, $b_1 + b_3 + b_5$, etc. Learning dynamics on $N$-1 RBM is in such way: on surface, it is dynamics on $\mathbb{R}^N$, but real learning dynamics is on those  $2^N 1.5^{2^N-N-1}$ regions (or X-forms). 
\bigskip

{\bf $N$-$M$ RBM} \\
Suppose $\mathcal{R}_i, i=1, \ldots, M$ are $M$ $N$-1 RBMs, we can form a $N$-$M$ RBM, denote as $\mathcal{R} = (\mathcal{R}_1, \ldots. \mathcal{R}_M)$, whose processing are  $p = (p_1, p_2, \ldots, p_M)$, where $p_i, i = 1, \ldots, M$ are processing of $\mathcal{R}_i$. So, $\mathcal{R}$ is Cartesian product of $\mathcal{R}_i, i=1, \ldots, M$. 

Since all $\mathcal{R}_i$ are cut into regions, and in each region, processing is same, we can see $\mathcal{R}$ is also cut into regions, and each region is a Cartesian product of regions of $\mathcal{R}_i$: $R = R_1 \times R_2 \times \ldots \times R_M$, where $R_i$ is one region from i-th RBM $\mathcal{R}_i$. Thus, the number of all possible regions of $\mathcal{R}$ is $(2^N 1.5^{2^N-N-1})^M = 2^{NM} 1.5^{M(2^N-N-1)}$. This is a much smaller number than $2^{M2^N}$, which is the number of total possible processing for $N$-$M$ IPU. 

X-form for each region of $\mathcal{R}$, are actually Cartesian product of X-form for those regions of $\mathcal{R}_i$. Suppose $R = R_1 \times R_2 \times \ldots \times R_M$, and $f_i$ are X-forms for region $R_i$ in $\mathcal{R}_i$, $i=1, \ldots, M$, then X-form for $R$ is $f = (f_1, \ldots, f_M)$.  For example, $(b_1, b_1+b_3, \ldots, b_2 \cdot b_4)$ is one X-form of $\mathcal{R}$. 

Learning dynamics on $N$-$M$ RBM is in such way: on surface, it is dynamics on parameter space $\mathbb{R}^NM$, but real learning dynamics is on those  $2^{NM} 1.5^{M(2^N-N-1)}$ regions (or X-forms). 
\bigskip

{\bf Stacking RBMs} \\
Consider a $N$-$M$ RBM $\mathcal{R}_1$, and a $M$-$L$ RBM $\mathcal{R}_2$, stacking them together, we get one $N$-$L$ IPU $\mathcal{R}$: A processing $p$ of $\mathcal{R}$ are composition of processing $p_1, p_2$ of $\mathcal{R}_1, \mathcal{R}_2$: $p(i_1, i_2, \ldots, i_N) = p_2 ( p_1(i_1, i_2, \ldots, i_N) )$. And we denote as: $\mathcal{R} = \mathcal{R}_1 \otimes \mathcal{R}$.

The parameter space of $\mathcal{R}$ clearly is $\mathbb{R}^{NM} \times \mathbb{R}^{ML}$. We know $\mathbb{R}^{NM}$ is cut into some regions, in each region processing is same. So does $\mathbb{R}^{ML}$. Thus, $\mathbb{R}^{NM + ML}$ is cut into some regions, in each region processing is same, and these regions are Cartesian product of regions in $\mathbb{R}^{NM}$ and $\mathbb{R}^{ML}$. So, we know number of total possible processing $\mathcal{R}$ equals total possible processing of $\mathcal{R}_1$ times $\mathcal{R}_2$, i.e. 
$2^{NM} 1.5^{M(2^N-N-1)} \times 2^{ML} 1.5^{L(2^M-M-1)} = 2^{NM+ML} 1.5^{M(2^N-N-1) + L(2^M-M-1)}$. 

We can easily see if $M$ is large enough, the above number will become greater than $2^{L2^N}$, which is total possible processing of $\mathcal{R}$. We can see, at least potentially, $\mathcal{R}$ has enough ability to become a $N$-$L$ full IPU. But, we will not consider here. In fact, it is very closely related to the so called Universal Approximation Theorem. Indeed, stacking RBM together is powerful. 

X-form can be expressed as composition as well. For example, consider 3 2-1 RBM $\mathcal{R}_1$, $\mathcal{R}_2$, and $\mathcal{R}_3$. Using $\mathcal{R}_1$ and $\mathcal{R}_2$ to form a 2-2 RBM, and using $\mathcal{R}_3$ to stack on it, we get a 2-1 IPU $\mathcal{R}$: $\mathcal{R} = \mathcal{R}_3 \otimes (\mathcal{R}_1, \mathcal{R}_2)$. If for this case, $\mathcal{R}_1$ has X-form $b_1$, and $\mathcal{R}_2$ has X-form $b_2$, and $\mathcal{R}_3$ has X-form $b_1 + b_2 + b_3$, them, $\mathcal{R}$ has X-form $(b_1 + b_2 + b_3) (b_1, b_2)$. Easy to see this X-form is processing $P_5$ (XOR gate), which is not expressible by one 2-1 RBM. So, putting 3 2-1 RBMs together, more X-form can be expressed.
\bigskip

{\bf Deep Learning and Learning Dynamics} \\
Following idea of "putting RBM together, then more expressible", by stacking more RBMs, deep learning is formed. Suppose $\mathcal{R}_j$, are $N_j$-$N_{j+1}$ RBM, $j = 1, \ldots, J$, where $N_1, \ldots, N_J, N_{J+1}$ are sequence of integers.  So, we can stack these RBM together to form one $N_1$-$N_{J+1}$ IPU, which has processing: $p(i_1, i_2, \ldots, i_{N_1}) = p_J ( \ldots  p_2 ( p_1(i_1, i_2, \ldots, i_{N_1}) )$
where each $p_j$ is processing of $\mathcal{R}_j$. We denote this IPU as $\mathcal{R} =  \mathcal{R}_1 \otimes \ldots \otimes \mathcal{R}_J$.

Deep learning is conduct on IPU $\mathcal{R}$. How is learning dynamics? As we have seen from simplest RBM to deep learning, we know that the parameter space is cut into regions, and each region is for one processing, and learning is moving from one region to another. In another words, on surface, learning dynamics is going on the huge parameter space $\mathcal{R}^{N_1N_2 + \ldots + N_JN_{J+1}}$, but is actually on those regions. The number of regions are huge: $2^{N_1N_2} 1.5^{N_2(2^{N_1}-N_1-1)} \times \ldots$. 

Thus, as learning, huge number of parameters are adapting. But only when parameters cross region, processing changes. Before crossing, processing remains same, parameter changes at most can be thought as the preparation for crossing (perhaps many parameter changes are just wasted). We can say, learning dynamics is to move from one region to another. We also know each region is represented by one X-form, thus, learning dynamics is to move from one X-form to another X-form.

Deep learning structure is formed by human intervene. Once it is formed, if no further human intervene, it is mechanical learning. But, once it is formed, the structure, such as how many RBMs, how to stack, how to do Cartesian product, etc, are fixed. Thus, regions are fixed. Thus, available X-forms for learning are fixed. Then learning is conducted on the available X-forms. In other words, learning dynamics is to find a good X-form among those available X-forms. There are quite a variety methods of learning, i.e. how to find the good X-form. But, most essential methods are 2: contrastive divergence (CD), and stochastic gradient descendant (SGD).

We can see one example. $\mathcal{R}_1$, $\mathcal{R}_2$, $\mathcal{R}_3$ are 3 2-1 RBMs. We put them like this: $\mathcal{R} = \mathcal{R}_3 \otimes (\mathcal{R}_1, \mathcal{R}_2)$. We have 3 parameter space $(a, b), (c, d), (e, f)$. We have 6 regions for each parameter space. Put them together, we have 6x6x6 = 216 regions. $\mathcal{R}$ is one 2-1 IPU. So, $\mathcal{R}$ at most has 8 processing. Thus, among those 216 regions some different regions will have same processing. But, each region will have one X-form. That is to say, for one processing, there could have several X-form associate with it. For example, consider this region: $R^3_1 \times (R^1_3, R^2_5)$. This gives processing $P_5$ (XOR gate). Normally, for this processing, we can use X-form $b_1 + b_2$ for it. But, for the region, naturally, the X-form is: $(b_1 + b_2 + b_3) \otimes (b_1, b_2)$. That is to say, this X-form will generate the same processing as $b_1 + b_2$. Another region: $R^3_2 \times (R^1_2, R^2_6)$ will give the same processing. And, the X-form is: $(b_1 + b_3) \otimes (b_1+b_3, b_2+b_3)$.

So, for deep learning, we can say that all X-forms form the internal representation space. But, for deep learning, this internal representation space could not be seen directly, it is embedded into huge parameter space   $\mathbb{R}^{N_1 N_2 + \ldots + N_J N_{J+1}}$, each X-form equivalent to one region in this parameter space, which is a Cartesian product of regions of $\mathcal{R}_j, j=1, \ldots, J$. 

So, we can see deep learning is doing: 1) to form a set of available X-forms and embed those X-forms into parameter space, this is done at the time to set up the learning structure and this is done by human; 2) to use some learning methods (e.g. CD+SGD) to form dynamics on parameter space, and use a big amount of data to drive the dynamics, 3) hope to reach a good X-form among the set. This is what we described in section 5, Strategy 1. In fact, the loss function used in SGD is exactly the function in the equation \eqref{eq:embk} in section 5. This is what happens inside a deep learning. 

Now, we have a clear picture on what deep learning is doing and why it is working well: It conducts learning by using Strategy 1 -- Embed X-forms into Parameter Space. Also, the data sufficiency that we defined actually lay down a framework for understanding what kind of data and what amount of data are sufficient to drive a successful learning. This is very important. We will no longer be blind about big data. We can calculate if data is sufficient for one task. This a huge contribution from our descriptions on mechanical learning to deep learning.
\bigskip

{\bf Disadvantages of Deep Learning} \\
Finally, we want to point out: deep learning might not be best mechanical learning. We list some its disadvantages below: 
\begin{enumerate} [topsep=0pt,itemsep=-1ex,partopsep=1ex,parsep=1ex]
\item It is acting on huge parameter space, but, actual learning dynamics is on a fixed set of regions (which is equivalent to a set of X-forms). This indirectness makes every aspects of learning harder, especially, it is near impossible to know what is exactly happening in learning dynamics.
\item As theorem 7 in section 5 shows, successful learning needs data sufficient to support and to bound. This requirement is very costly.
\item The structure of learning is setup by human. Once setup, structure (how many layers, how big a layer is, how layers fitting together, how to do convolution, etc) is not be able to change. This means that learning is restricted on a fixed group of regions, equivalent a fixed group of X-forms. If best X-form is not in this set, deep learning has no way to reach best X-form, no matter how big data are and how hard to try. Consequently, it is not universal learning machine.
\item It is very costly to embed X-forms into a huge parameter space. Perhaps, among all computing spend on learning, only a very small fraction is used on critical part, i.e. on moving X-form, and most are just wasted.
\item Since there is no clear and reachable internal representation (due to the embedding), it will be very hard to do advanced learning, such as to unite all 5 learning methods together (see \cite{pedro}).
\item Since there is no clear internal representation space, it is hard to define initial X-form, which is very essential to efficiency improving and several stages of learning.
\end{enumerate}
Knowing the disadvantages is good, we can find some better way to turn disadvantages to advantages. 
\bigskip

\end{document}